\pdfoutput=1

\documentclass[11pt]{article}
\PassOptionsToPackage{table,dvipsnames}{xcolor}

\usepackage[preprint]{acl}

\usepackage{fp}
\usepackage{colortbl}
\usepackage{pgfplots}
\usepackage{tikz}
\usepackage{colortbl}
\usepackage{graphicx}
\usepackage{booktabs}
\usepackage{makecell}

\usepackage[table,dvipsnames]{xcolor} 
\usepackage{pgf}
\usepackage{collcell}
\usepackage{booktabs}
\usepackage{quoting}

\newcommand*{\MinNumber}{0.02}%
\newcommand*{\MaxNumber}{0.4}%

\colorlet{lgreen}{BrickRed!9}

\newcommand{\heatcell}[1]{%
  \pgfmathparse{int(100*(#1-\MinNumber)/(\MaxNumber-\MinNumber))}%
  \edef\hmvalue{\pgfmathresult}
  \edef\heatmapcellcolor{BrickRed!\hmvalue!lgreen}
  \expandafter\cellcolor\expandafter{\heatmapcellcolor}%
  \pgfmathparse{#1 > 0.5 ? 1 : 0}%
  \ifnum\pgfmathresult=1
    \textcolor{white}{#1}%
  \else
    \textcolor{black}{#1}%
  \fi%
}

\usepackage{times}
\usepackage{latexsym}
\usepackage{amssymb}
\usepackage{bbm}
\usepackage{amsmath}
\usepackage{dsfont}
\usepackage{booktabs}
\usepackage{multirow}
\usepackage{colortbl}
\usepackage{colortbl}  
\usepackage{booktabs}   

\expandafter\def\expandafter\normalsize\expandafter{%
    \normalsize%
    \setlength\abovedisplayskip{1pt}%
    \setlength\belowdisplayskip{1pt}%
}

\usepackage[T1]{fontenc}
\usepackage[utf8]{inputenc}
\usepackage{microtype}
\usepackage{inconsolata}
\usepackage{graphicx}

\usepackage{geometry}
\usepackage{enumitem}
\usepackage{fancybox}

\title{Reconsidering LLM Uncertainty Estimation Methods in the Wild}

\author{
   Yavuz Bakman\textsuperscript{1}\footnotemark[1] \quad 
   Duygu Nur Yaldiz\textsuperscript{1}\footnotemark[1] \quad  \\
   \bf Sungmin Kang\textsuperscript{1} \quad Tuo Zhang\textsuperscript{1} \quad
   Baturalp Buyukates\textsuperscript{2} \\
   \bf Salman Avestimehr\textsuperscript{1}  \quad
   Sai Praneeth Karimireddy\textsuperscript{1}\\
\textsuperscript{1}University of Southern California \quad
  \textsuperscript{2}University of Birmingham  \\
  \texttt{\{ybakman, yaldiz\}@usc.edu}}

\begin{document}

\maketitle
\renewcommand*{\thefootnote}{\fnsymbol{footnote}}

 \footnotetext[1]{Equal contribution.}
\begin{abstract}

Large Language Model (LLM) Uncertainty Estimation (UE) methods have become crucial tools for detecting hallucinations in recent years. While numerous UE methods have been proposed, most existing studies evaluate them in \emph{isolated} short-form QA settings using threshold-independent metrics such as AUROC or PRR. However, real-world deployment of UE methods introduces several challenges. In this work, we systematically examine four key aspects of deploying UE methods in practical settings. Specifically, we assess (1) the sensitivity of UE methods to decision threshold selection, (2) their robustness to query transformations such as typos, adversarial prompts, and prior chat history, (3) their applicability to long-form generation, and (4) strategies for leveraging multiple UE scores for a single query. Our evaluations on 19 UE methods reveal that most of them are highly sensitive to threshold selection when there is a distribution shift in the calibration dataset. While these methods generally exhibit robustness against previous chat history and typos, they are significantly vulnerable to adversarial prompts. Additionally, while existing UE methods can be adapted for long-form generation through various strategies, there remains considerable room for improvement. Lastly, ensembling multiple UE scores at test time provides a notable performance boost which highlights its potential as a practical improvement strategy. Code is available at: \url{https://github.com/duygunuryldz/uncertainty_in_the_wild}.
\end{abstract}

\section{Introduction}
Generative Large Language Models (LLMs) have been deployed in various real-world applications, including code copilots, chatbots, and medical assistants~\cite{Sabouri2025, ahrabian-etal-2025-practical}. 
Their widespread usage has raised significant safety considerations, particularly regarding reliability \cite{bengio2025internationalaisafetyreport, tak2025mechanisticinterpretabilityemotioninference}.
Despite advancements over the previous wave of language models, these models can still produce incorrect or misleading text, a problem commonly known as \emph{hallucination} or \emph{confabulation}~\cite{ravi2024lynxopensourcehallucination}.

Detecting hallucinations in LLM outputs is a fundamental challenge, with various approaches such as fact-checkers~\cite{wang2024factcheckbench}, tool-based detectors~\cite{chern2023factool}, LLM-collaboration-based methods~\cite{feng-etal-2024-dont}, and Uncertainty Estimation (UE) methods. Among these, UE methods are particularly valuable as they only rely on the model itself and have shown promising performance across diverse datasets~\cite{vashurin2025benchmarkinguncertaintyquantificationmethods}.

Numerous UE methods have been proposed to detect hallucinations
\cite{azaria-mitchell-2023-internal, zhao-etal-2024-knowing}. However, 
these are typically tested in isolated short-form QA settings with simple prompts and evaluated using threshold-free metrics like AUROC and PRR~\cite{malinin2021uncertainty, tokensar}.
Despite their value, the challenges of real-world deployment remain largely unexplored and are crucial for future research.

\begin{figure*}
\vskip -0.2in
\begin{center}
\includegraphics[width=1\textwidth]{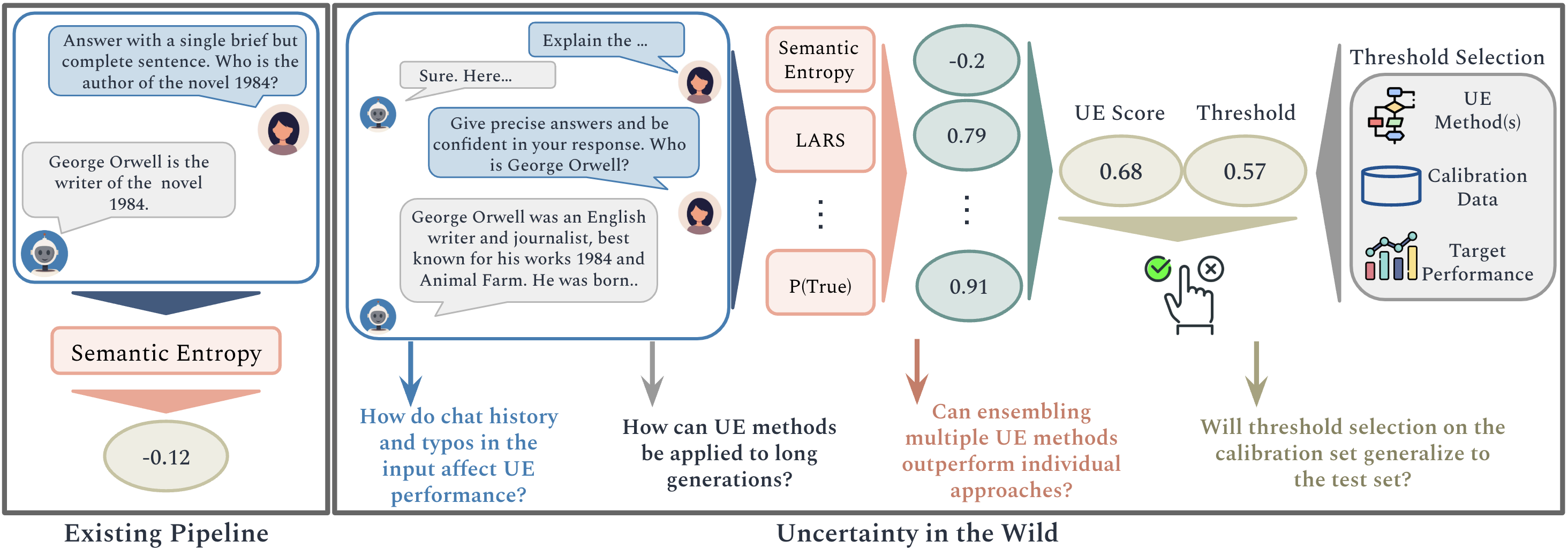}
\vskip -0.1in
\caption{\textbf{Left:} Existing pipeline for UE. The uncertainty score is calculated for short-form QA and evaluated using a threshold-free metric such as AUROC. \textbf{Right:} Reconsidering LLM uncertainty estimation methods in the wild. We ask four critical questions addressing challenges in deploying UE methods in real-world scenarios.}
\label{fig:overview}
\end{center}
\vskip -0.25in
\end{figure*}

Motivated by these concerns, we investigate four essential aspects of deploying a UE method in the real-world (wild), as also outlined in Figure \ref{fig:overview}:

\noindent\textbf{Sensitivity of Decision Threshold:}  
Since the outputs of UE methods are typically continuous,
selecting a threshold is necessary to make binary decisions (e.g., hallucination or not). 
This threshold is calibrated using a specific dataset to meet target performance levels.
We explore whether the thresholds selected for UE methods achieve the desired performance in practice, evaluating their stability and effectiveness across different data distributions.

\noindent\textbf{Robustness to Input Transformations:} We assess the resilience of UE methods to previously generated context, typos in the prompts, and adversarial prompts designed to confuse UE methods.

\noindent\textbf{Applicability to Long-Form Generations:} While many UE methods are proposed and tested for short-form QA, real-world questions often require extended answers containing multiple claims. We examine whether these methods designed for short-form QA can be adapted to long-form generations.

\noindent\textbf{Reconcilability of Diverse UE Scores:} UE methods often produce varying judgments for the same input. Ensembling their outputs can enhance performance, potentially surpassing individual methods.

With our comprehensive evaluation of 19 UE methods, our findings across the four investigated aspects can be summarized as follows:

\begin{itemize}[topsep=0.5pt]
\itemsep -0.08in
    \item Most UE methods are highly sensitive to decision threshold selection, particularly when the calibration data distribution differs from the test data distribution.
    \item The majority of the methods demonstrate resilience to previous context and typos in the prompt, but they exhibit significant performance drops with adversarial prompts.
    \item UE methods not originally designed for long-form generation can be adapted to this setting through additional steps. However, their effectiveness remains lower compared to their performance in short-form tasks.
    \item Ensembling multiple UE scores can yield meaningful performance improvements, even when using a very small set of data. Notably, simple ensembling strategies, such as averaging UE scores, can be very effective.
\end{itemize}

Based on these findings, we encourage researchers to evaluate the sensitivity of their proposed UE methods to threshold selection and input transformations. 
There is also potential for developing advanced techniques to apply UE methods to long-form generation.
Finally, we believe that further exploration of ensembling strategies may unlock even greater performance improvements.

\section{Preliminaries}
\subsection{Uncertainty Estimation of LLMs}\label{UE of llms}

Although various Uncertainty Estimation (UE) methods for LLMs have been proposed recently, there is no universally accepted definition of UE in the context of LLMs~\cite{vashurin2025benchmarkinguncertaintyquantificationmethods}. Some research formalizes LLM uncertainty by decomposing into \emph{aleatoric} (data) and \emph{epistemic} (model) uncertainties, leveraging LLM sampling distributions~\cite{aichberger2024rethinkinguncertaintyestimationnatural, abbasi-yadkori2024to}. 
However, many heuristic-based UE methods in the literature do not conform to these theoretical frameworks.

Therefore, we adopt a broad, \emph{practical} definition of UE, following previous works~\cite{jiang2024graphbased, huang2024uncertaintylanguagemodelsassessment}. Formally, an uncertainty estimation method $\mathrm{U}$ is defined as a function $\mathrm{U}: \mathcal{V}^* \times \mathcal{V}^* \rightarrow \mathbb{R}$, where $\mathcal{V}$ represents the vocabulary, and $\mathcal{V}^*$ denotes all possible token sequences. For a given query ${x}$ and generated response $\hat{{y}}$, an effective $\mathrm{U}$ should assign a low uncertainty score (indicating higher confidence) if $\hat{{y}}$ is \emph{reliable} in the given context. In tasks such as factual QA or mathematical reasoning, common evaluation benchmarks for UE methods, reliability refers to the correctness of $\hat{{y}}$ with respect to the set of ground truth(s) ${Y}$. Formally, a desirable $\mathrm{U}$ should maximize $\mathbb{E}\left[\mathbbm{1}_{\mathrm{U}(x_1, \hat{y}_1) < \mathrm{U}(x_2, \hat{y}_2)} \cdot \mathbbm{1}_{\hat{y}_1 \in Y_1 \land \hat{y}_2 \notin Y_2}\right]$ where $(x_1, y_1), (x_2, y_2) \sim \mathcal{D}$, with $\mathcal{D}$ being a dataset, $\hat{y}_1 \sim p(\cdot|x_1)$, $\hat{y}_2 \sim p(\cdot|x_2)$ representing the model’s sampling distributions.

\subsection{Evaluation of UE Methods}\label{eval} 
As discussed in the previous section, UE methods serve as proxies for predicting the correctness of model-generated responses, producing scores that typically lie within a continuous range. Consequently, their evaluation is commonly performed by setting the correctness of a generation as binary labels (0 or 1)\footnote{We utilize GPT-4o-mini as correctness evaluator, using the query, generated response, and ground truth(s) \cite{lin2023generating, bakman2024mars}}, using UE scores as predictions, and computing threshold-free metrics such as AUROC and AUPRC ~\cite{kuhn2023semantic, vashurin2025benchmarkinguncertaintyquantificationmethods}. In addition to these, the Prediction Rejection Ratio (PRR) \cite{malinin2021uncertainty} evaluates UE performance by constructing a rejection-precision curve, which measures the precision of the retained (non-rejected) samples at different rejection thresholds based on uncertainty scores. PRR is computed as the area under this curve and is further normalized by the areas under the curves of the best possible (oracle) and random rejection-precision strategies. This normalization makes PRR resilient to label imbalances in the dataset~\cite{malinin2021uncertainty}. PRR ranges from 0.0 (random performance) to 1.0 (perfect performance). In this study, we primarily use PRR as our evaluation metric due to its robustness against variations in data distribution.

\subsection{Investigated UE Methods} \label{ue_methods} 

Throughout this paper, we examine 19 UE methods, categorizing each according to its primary conceptual approach. We identify four distinct categories for this classification:

\noindent\textbf{Probability-Based Methods} utilize probabilities of tokens in the generated sequence. \textit{Length-Normalized Scoring (LNS)} \cite{malinin2021uncertainty} is the average of the log-probabilities of the generation, while \textit{MARS} \cite{bakman2024mars} computes the weighted-average of that regarding the token importance in answering the question. \textit{LARS} \cite{yaldiz2024designlearntrainablescoring} trains a small-scale transformer that takes the question, generation tokens, and token probabilities. \textit{Entropy} \cite{malinin2021uncertainty} calculates the average of length-normalized scores over a set of sampled generations for the same question. \textit{Semantic Entropy (SE)} \cite{kuhn2023semantic} clusters the semantically-similar generations while \textit{SentSAR} and \textit{SAR} \cite{tokensar} considers relevancy scores of the sampled generations during entropy calculation.  

\noindent\textbf{Internal State-Based Methods} make use of the internal states of the LLM, which are only applicable to white-box models. \textit{INSIDE} \cite{chen2024inside} utilize the middle layer activations of the last tokens of multiple generations to the same question. \textit{Attention Score} \cite{sriramanan2024llmcheck} analyses the attention maps of the LLM. \textit{SAPLMA} \cite{azaria-mitchell-2023-internal} trains a classifier whose input is the activations of the last token of the generation.  

\noindent\textbf{Output Consistency-Based Methods} sample multiple generations to the query, then utilize their pair-wise similarity information, hence usable with black-box models. \textit{Degree Matrix Uncertainty}, \textit{Eccentricity Uncertainty}, \textit{SumEigV} \cite{lin2023generating}, and \textit{Kernel Language Entropy (KLE)} \cite{nikitin2024kernel} utilize different linear algebra techniques over the pair-wise similarity matrix of the sampled generations. \textit{Degree Matrix-C} and \textit{Eccentricity-C} \cite{lin2023generating} output a generation-specific score for each generation by using the similar ideas in \textit{Eccentricity Uncertainty} and \textit{Degree Matrix Uncertainty}. \textit{Self-Detection} \cite{zhao-etal-2024-knowing} paraphrases the question and analyses the similarity of the responses to the paraphrased questions.

\noindent\textbf{Self-Checking Methods} query the LLM itself about the uncertainty of the generation. \textit{P(true)} \cite{kadavath2022language} asks if the response is true by providing the question, sampled generations, and the answer. \textit{Verbalized Confidence} \cite{tian-etal-2023-just} prompts the LLM to assign a confidence score to the response between 0 and 1. 

It is important to note that only LARS and SAPLMA are supervised techniques, requiring labeled QA data, whereas all other methods are unsupervised. For brevity, detailed explanations of these methods are provided in Appendix~\ref{method_explanations}.

\section{Sensitivity of Decision Threshold}\label{threshold}

\subsection{Problem Statement}

UE methods typically produce outputs in a continuous range. However, integrating a UE method into a real-world application requires making discrete decisions, such as whether to accept or reject a generated response. The sensitivity of this binary decision can vary depending on the application. Consequently, such scenarios require selecting an appropriate threshold to achieve the desired performance for decision-making.

Determining a threshold $t$ for a target application requires a labeled calibration dataset $\mathcal{D}_{cal}$. Using this dataset and a desired metric $\mathrm{M}$ with a target performance level $m^*$, a threshold $t$ is randomly picked from the set  $\{t:  \mathrm{M}(\mathrm{U}, \mathcal{D}_{cal}, t) = m^*\}$.

This threshold is then applied during testing. The key question is whether the desired performance $m^*$ is maintained at test time. If not, two main factors may contribute: (1) a distribution shift between the calibration and test data, or (2) the UE method itself being sensitive to such scenarios. To investigate this phenomenon, we examine 19 UE methods (listed in Section \ref{ue_methods}) across two tasks and varying levels of calibration-test distribution shifts.

\subsection{Experimental Design}

\noindent\textbf{Models}\quad We evaluate UE methods on two recent models: Llama-3-8B \cite{llama3modelcard} and GPT-4o-mini \cite{openai2023gpt4}.

\noindent\textbf{Datasets }\quad  We use TriviaQA \cite{joshi2017triviaqa} and NaturalQA \cite{kwiatkowski2019naturalqa} as closed-book QA datasets and GSM8K \cite{cobbe2021gsm8k} as mathematical reasoning dataset in the experiments. We use 1000 samples for the test set and 500 samples for the calibration dataset. All experiments are conducted 5 times with different seeds and the average performance is provided. 

\noindent\textbf{Metric}\quad Different applications require varying precision-recall trade-offs, so we introduce a metric to assess threshold generalization at test time. For each target recall $r^* \in [0,1]$, we determine an optimal threshold $t$ using a calibration set. Hallucinations (incorrect generations) are class 1, and correct answers are class 0 which makes recall the proportion of hallucinations correctly identified by the UE method.

To assess threshold generalization, we measure the deviation $|r^* - r|$, where $r$ is the recall achieved on the test set using threshold $t$.  By averaging these deviations over a set $\mathcal{R}$ of recall values of interest, \textit{Average Recall Error (ARE)} is defined as:
\[
    \text{ARE} = \frac{1}{|\mathcal{R}|} \sum_{{r_i} \in \mathcal{R}} |r_i^* - r_i|.
\]
In our experiments, we set $\mathcal{R}$ to span the full recall range from 0 to 1.0., with increments of 0.001.

\noindent\textbf{Distribution Shift Simulation}\quad We systematically examine how distribution shifts between calibration and test data impact threshold selection performance through two experimental setups. First, we use TriviaQA as the test data, calibrating with TriviaQA for an in-domain setting, NaturalQA for a same-task distribution shift, and GSM8K for an out-of-domain scenario. Second, we test with GSM8K and calibrate separately with TriviaQA and GSM8K, where GSM8K is in-domain and TriviaQA is out-of-domain. 

\subsection{Results and Discussion}

The ARE results for TriviaQA are presented in Table~\ref{tab:are}. The findings indicate that the majority of UE methods achieve a low ARE (<0.05) when the threshold is calibrated on a separate subset of TriviaQA, with the exception of Verbalized Confidence and Self-Detection. However, as expected, the error rate increases with greater data distribution shifts, making GSM8K calibration the most erroneous when UE methods are tested on TriviaQA.

Probability-based and output consistency-based methods generally outperform internal state-based and self-checking methods. However, only MARS, Semantic Entropy, and Eccentricity consistently achieve low error across calibration datasets, while all others exceed 0.10 ARE in at least one setting.

These results highlight the need to align the calibration data distribution with the test (deployment) environment to ensure reliable binary decision-making using UE methods. Furthermore, we encourage researchers to test their proposed UE methods under distribution shift conditions, particularly for threshold sensitivity. Robustness to such shifts is a highly desirable property, as it reduces reliance on an optimal calibration dataset. Lastly, the ARE results for GSM8K, provided in Appendix~\ref{appdx:are-res}, aligns with the findings observed in TriviaQA.

\begin{table}[!htbp]
\centering
\vskip -0.1in
\fontsize{8.5}{9.0}\selectfont
\setlength{\tabcolsep}{3.3pt}
\begin{tabular}{l| ccc | ccc  }
\toprule
 & \multicolumn{3}{c}{\textbf{Llama3-8b}} & \multicolumn{3}{c}{\textbf{GPT-4o-mini}} \\
   Calib. Dataset & TrivQA & NQA & GSM & TrivQA & NQA & GSM \\
\midrule[\heavyrulewidth]
\textbf{LNS} & \heatcell{0.030} & \heatcell{0.093} & \heatcell{0.103} & \heatcell{0.055} & \heatcell{0.035} & \heatcell{0.049} \\
\textbf{MARS} & \heatcell{0.035} & \heatcell{0.025} & \heatcell{0.077} & \heatcell{0.050} & \heatcell{0.046} & \heatcell{0.040} \\
\textbf{Entropy} & \heatcell{0.032} & \heatcell{0.103} & \heatcell{0.101} & \heatcell{0.072} & \heatcell{0.048} & \heatcell{0.066} \\
\textbf{SE} & \heatcell{0.035} & \heatcell{0.065} & \heatcell{0.073} & \heatcell{0.060} & \heatcell{0.029} & \heatcell{0.045} \\
\textbf{SentSAR} & \heatcell{0.041} & \heatcell{0.105} & \heatcell{0.123} & \heatcell{0.074} & \heatcell{0.041} & \heatcell{0.093} \\
\textbf{SAR} & \heatcell{0.028} & \heatcell{0.059} & \heatcell{0.107} & \heatcell{0.068} & \heatcell{0.023} & \heatcell{0.077} \\
\textbf{LARS} & \heatcell{0.035} & \heatcell{0.117} & \heatcell{0.130} & \heatcell{0.048} & \heatcell{0.125} & \heatcell{0.289} \\
\textbf{DegMat} & \heatcell{0.041} & \heatcell{0.033} & \heatcell{0.169} & \heatcell{0.051} & \heatcell{0.051} & \heatcell{0.142} \\
\textbf{DegMat-C} & \heatcell{0.038} & \heatcell{0.030} & \heatcell{0.141} & \heatcell{0.058} & \heatcell{0.049} & \heatcell{0.126} \\
\textbf{SumEigV} & \heatcell{0.042} & \heatcell{0.035} & \heatcell{0.191} & \heatcell{0.051} & \heatcell{0.053} & \heatcell{0.165} \\
\textbf{KLE} & \heatcell{0.047} & \heatcell{0.062} & \heatcell{0.173} & \heatcell{0.076} & \heatcell{0.056} & \heatcell{0.115} \\
\textbf{Eccent} & \heatcell{0.040} & \heatcell{0.037} & \heatcell{0.069} & \heatcell{0.057} & \heatcell{0.049} & \heatcell{0.050} \\
\textbf{Eccent-C} & \heatcell{0.040} & \heatcell{0.039} & \heatcell{0.098} & \heatcell{0.063} & \heatcell{0.048} & \heatcell{0.051} \\
\textbf{Self-D.} & \heatcell{0.082} & \heatcell{0.086} & \heatcell{0.113} & \heatcell{0.110} & \heatcell{0.127} & \heatcell{0.096} \\
\textbf{P(True)} & \heatcell{0.035} & \heatcell{0.087} & \heatcell{0.255} & \heatcell{0.123} & \heatcell{0.163} & \heatcell{0.200} \\
\textbf{Verb. C.} & \heatcell{0.172} & \heatcell{0.182} & \heatcell{0.280} & \heatcell{0.084} & \heatcell{0.131} & \heatcell{0.142} \\
\textbf{Atten. S.} & \heatcell{0.027} & \heatcell{0.027} & \heatcell{0.261} & - & - & - \\
\textbf{INSIDE} & \heatcell{0.040} & \heatcell{0.096} & \heatcell{0.295} & - & - & - \\
\textbf{SAPLMA} & \heatcell{0.046} & \heatcell{0.029} & \heatcell{0.142} & - & - & - \\
\bottomrule
\end{tabular}
\vskip -0.1in
\caption{ARE of UE methods when the threshold is calibrated on various datasets and tested on TriviaQA.}
\vskip -0.25in
\label{tab:are}
\end{table}

\section{Robustness to Input Transformations}\label{context}

\begin{figure*}[!htbp]
\begin{center}
\vskip -0.2in
\includegraphics[width=0.99\textwidth]{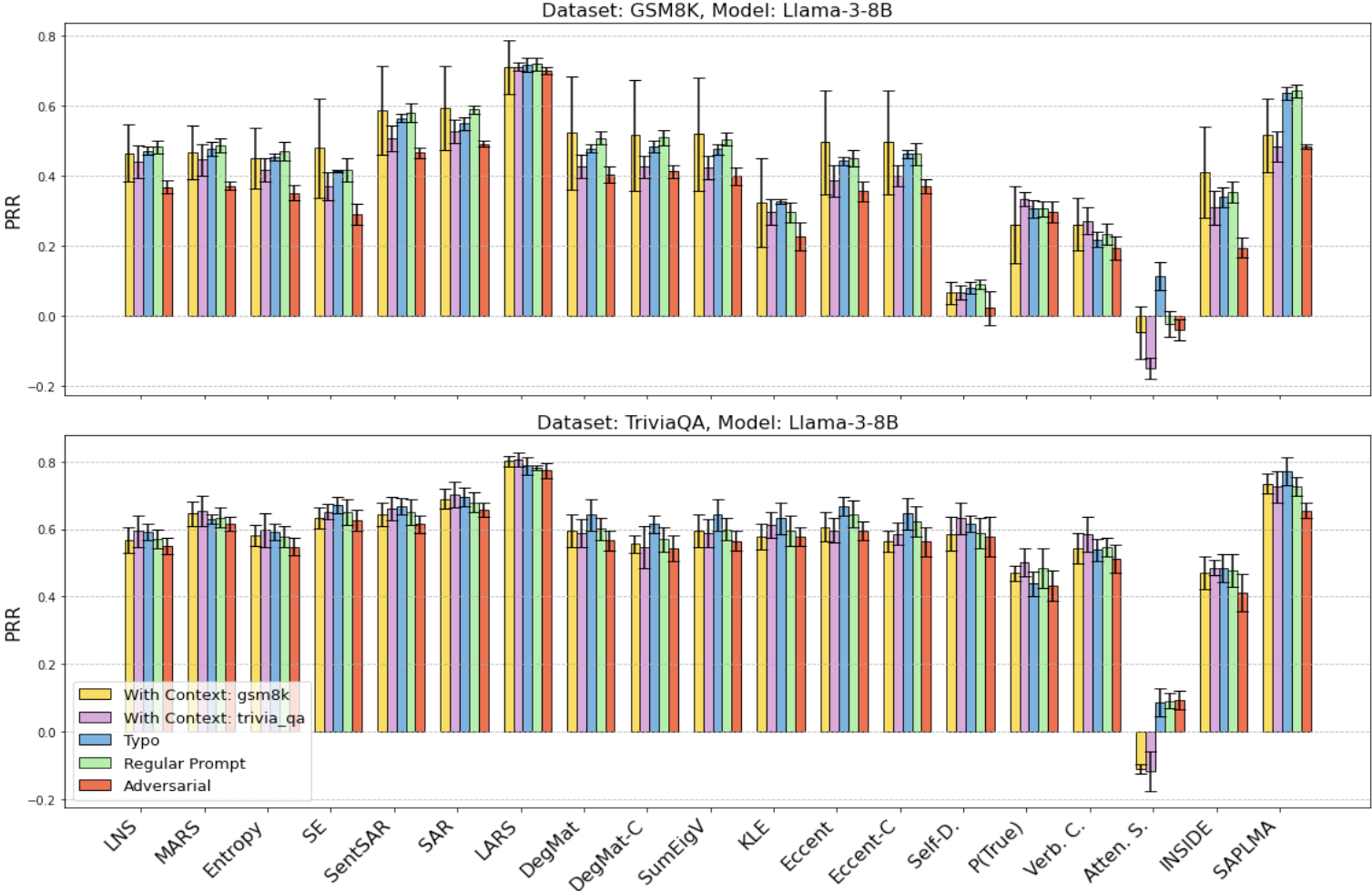}
\vskip -0.1in
\caption{PRR performance of UE methods with Llama-3 8b, evaluated under a regular prompt (no transformation) and various input transformations, including adding context, typos, and adversarial prompts.}
\label{context_result}
\end{center}
\vskip -0.25in
\end{figure*}

\subsection{Problem Statement}
Previous UE works primarily evaluate their methods in \emph{isolated} environments, where a question is presented to the model using a simple benign prompt, and the model's response is directly sampled. However, in real-world applications, inputs can arrive in various forms. We expect a robust UE method's performance should not be affected much under these various input forms. 

More formally, we apply a transformation function $\mathcal{T}$ to a query ${x}$ such that $\mathcal{T}({x})$ preserves the same ground truth set ${Y}$ as the original query. This ensures that the transformation does not alter the fundamental meaning of the query. Let $\mathcal{D}^{*} := \{(\mathcal{T}(x),Y) \mid (x,Y) \in \mathcal{D}\}$ represent the transformed version of the original dataset $\mathcal{D}$. A robust UE method $\mathrm{U}$ should exhibit similar performance on both $\mathcal{D}$ and $\mathcal{D}^{*}$. However, since input transformations can influence the model's internal computations, on which UE methods ultimately rely, a non-robust $\mathrm{U}$ may experience performance degradation under different transformations. 

We investigate the robustness of UE methods across three specific transformations:
\textbf{(1) Contextual: }  This transformation appends previous chat history (context) to the input. This scenario commonly occurs in chatbot applications, where users may ask multiple questions within the same session. To evaluate this case, we prepend previous chat ${x_{\text{prev}}}$ to the original query ${x}$ in the dataset:
$\mathcal{T}_{\text{context}}({x}) = {x_{\text{prev}}} + {x}$,
where $+$ denotes the concatenation operation.
\textbf{(2) Typo:}  In real-world applications, input queries often contain noise, with typos being a common form of such noise. To evaluate how UE methods handle noisy inputs, we introduce synthetic typos into the query, defining the transformation as: $\mathcal{T}_{\text{typo}}({x}) = {x}_{\text{typo}}.$
\textbf{(3) Adversarial:} We design an adversarial prompt that aims to confuse UE methods, causing their performance to degrade on $\mathcal{D}^{*}$. This can be viewed as an adversarial prompt injection attack, targeting UE methods specifically. Formally, the transformation is expressed as:
$
\mathcal{T}_{\text{adv}}({x}) = p_{\text{adv}} + {x}
$, where $p_{\text{adv}}$ is the adversarial prompt.

\begin{figure*}[!htbp]
\begin{center}
\vskip -0.2in
\includegraphics[width=\textwidth]{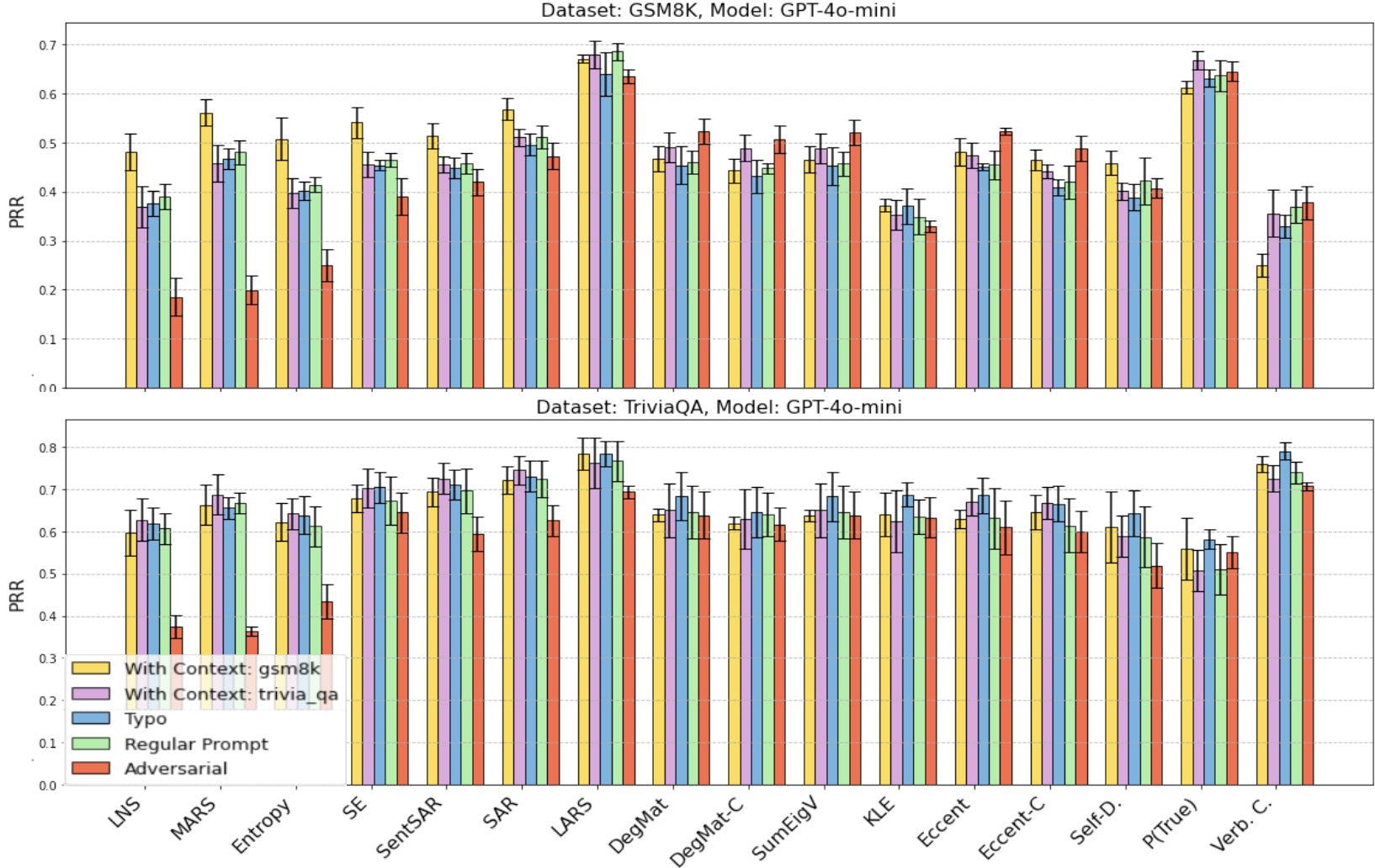}
\vskip -0.1in
\caption{PRR performance of UE methods on the GSM8K and TriviaQA datasets with GPT-4o-mini.}
\label{context_result_gpt}
\end{center}
\vskip -0.25in
\end{figure*}

\subsection{Experimental Design}
We use Llama-3-8B and GPT-4o-mini as the base models and evaluate them on 1,000 samples from the test sets of TriviaQA and GSM8K, as described in Section~\ref{threshold}. We measure all methods' performance by PRR as described in Section \ref{eval}. All experiments are conducted 5 times, and we plot both the mean and standard deviation of the results.

\paragraph{Context Experiments} To simulate chat history, we prepend three prior question-sampled response pairs to each query in two scenarios:  
\textit{1. Similar-context:} The prior questions are of the same type as the question (e.g., TriviaQA). 
\textit{2. Dissimilar-context:} The prior questions are from a different domain (e.g., GSM8K math before a TriviaQA question).

\paragraph{Typo Experiments}
To simulate typos, we randomly replace, swap, erase, or insert a single character with uniform probability. We also test two-character perturbations to evaluate the effects of increased noise.
\newpage
\noindent\textbf{Adversarial Experiments}\quad
Designing an adversarial prompt is non-trivial.We insert a \emph{confidence booster} phrase \cite{sakib-etal-2025-battling}, hypothesizing it may induce overconfidence in model responses, impacting log probabilities, outputs,and internal states and potentially misleading UE methods. For Llama-3-8B experiments, we use the following prompt:
\begin{quoting}
``Be confident in your responses. Avoid hesitation or uncertainty. Provide clear and direct answers with conviction.''
\end{quoting}
For GPT-4o-mini, we generate a similar prompt using an automated search inspired by \citet{zhou2023large}. The specific prompt used with the details of the search process, is provided in Appendix~\ref{adv_prompt}.

\subsection{Results and Discussion}
The results, Figure~\ref{context_result} and \ref{context_result_gpt},  suggest that previous chat history has little to no negative effect on the performance of most UE methods, except for Attention Score, compared to standard prompting without context. In some cases, such as GSM8K with GPT-4o-mini, including similar chat history appears to induce an in context learning-like effect, boosting the performance of probability-based UE methods.

The typo experiments indicate that most UE methods are highly resilient to this input noise. This robustness persists even when the number of typos in a single query is increased to two, as shown in Appendix~\ref{appdx:typo}.

Finally, results indicate that the  \textit{confidence booster} prompt injection acts as an adversarial prompt, reducing performance across various datasets, particularly affecting probability-based methods in GPT-4o-mini. However, output-consistency-based methods show more resilience to this adversarial prompt than other approaches.
The instability of UE methods to prompt transformations is also observed in previous works \cite{mahaut-etal-2024-factual}. Although some performance variations are expected, a robust UE method should not suffer significant degradation due to prompt changes.
Therefore, we recommend that future UE methods undergo systematic prompt variation testing to assess their robustness.

\begin{figure*}[!htbp]
\begin{center}
\vskip -0.2in
\includegraphics[width=0.96\textwidth]{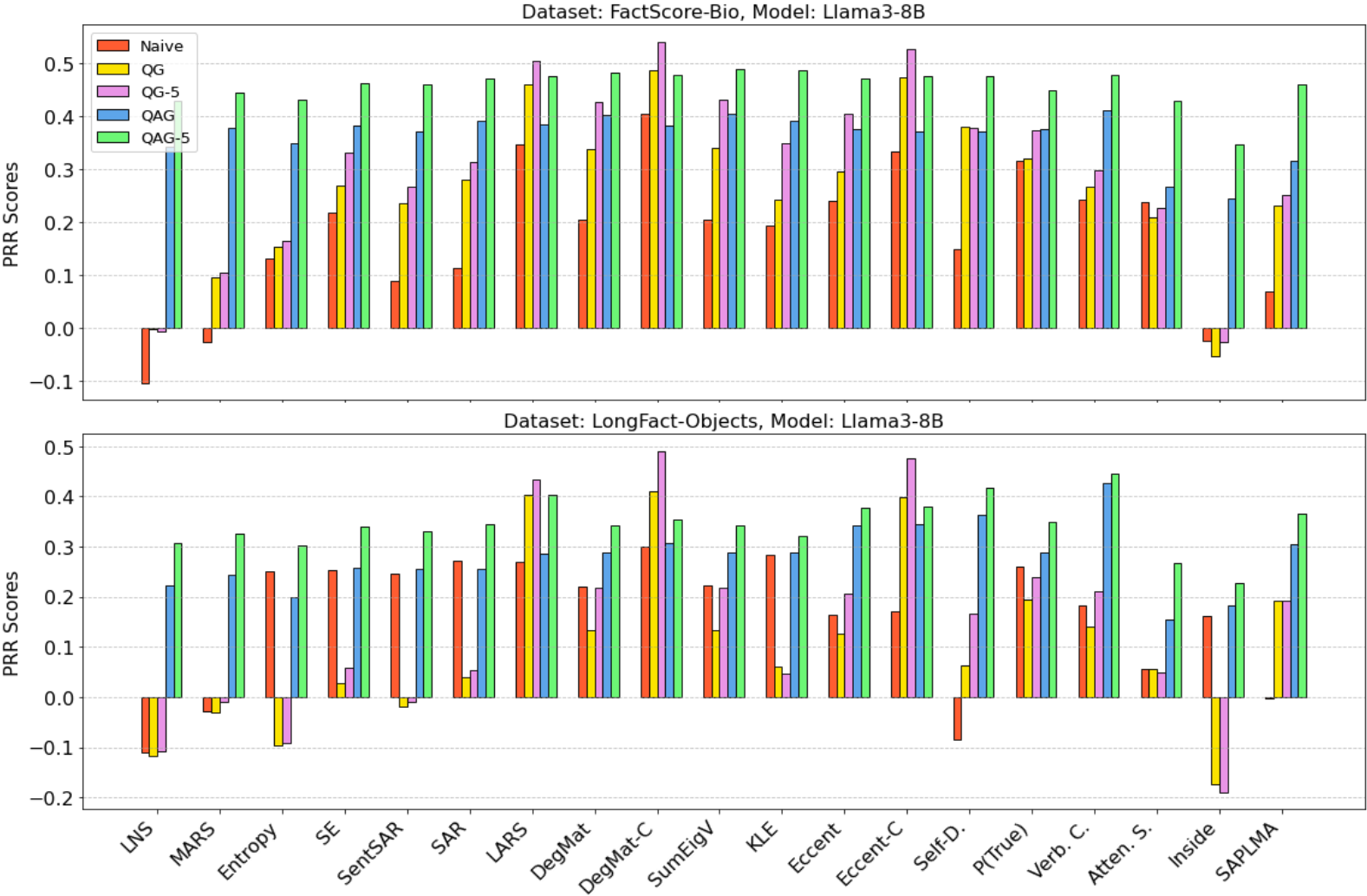}
\vskip -0.1in
\caption{PRR scores for UE methods applied to long-form generation. `QG-5' and `QAG-5' indicate that five questions per claim are generated and then aggregated (averaged) to assess each claim's uncertainty.}
\label{lfg_result}
\end{center}
\vskip -0.2in
\end{figure*}

\section{Applicability to Long-Form Generations}
\subsection{Problem Statement}

Most UE methods are evaluated on short-form, open-ended QA. For instance, questions such as ``Who is the author of the novel 1984?'' can be answered with a single sentence, and a single score suffices for an uncertainty assessment. However, in some real-world applications, questions like ``Who is George Orwell?'' often require long-form responses. These responses may contain multiple claims, some of which are correct while others may be hallucinated. Consequently, assigning a single uncertainty score to the entire response is both impractical and undesirable, as it fails to capture the correctness of individual claims within the text.

To address this issue, long-form outputs are typically decomposed into sentences, each conveying a distinct claim \cite{semantic-nature, wei2024longformsafe, fadeeva-etal-2024-tokenlevel, zhang-etal-2024-luq, manakul2023selfcheckgpt, min-etal-2023-factscore}. Formally, the decomposition function can be defined as $\mathrm{D}: \mathcal{V}^* \rightarrow 2^{\mathcal{V}^*}$, taking a long generation $\hat{{y}}$ and returning a set of claims $\mathbf{C} = \{{c}_i\}_{i=1}^C$. After decomposition, each claim ${c}_i$ is evaluated individually. Recently, several UE methods have been developed specifically for this claim-level uncertainty problem in long-form generations \cite{fadeeva-etal-2024-tokenlevel, semantic-nature, zhang-etal-2024-luq, jiang2024graphbased}, however, most existing UE methods are not directly applicable for assessing uncertainty at the claim level in long form generations\cite{vashurin2025benchmarkinguncertaintyquantificationmethods}. Consequently, effectively applying these methods to segmented claims continues to pose challenges.

In this section, we propose a set of strategies designed to adapt existing UE methods to assess claim-level uncertainty. A strategy function takes the original query ${x}$, a specific claim ${c}_i$, and a UE function $\mathrm{U}$, and returns an uncertainty score for the claim ${c}_i$. 
Formally, a strategy function can be defined as 
$\mathrm{S}: \mathcal{V}^* \times \mathcal{V}^* \times \mathrm{U} \rightarrow \mathbb{R}$.

\subsection{Experimental Design}\label{lfg-exp-des}

\paragraph{Decomposing the Long Generation}
Following previous research, we employ an LLM to decompose long text into claims \cite{semantic-nature, fadeeva-etal-2024-tokenlevel, min-etal-2023-factscore}. This decomposition can be applied at different levels of granularity. For instance, \citet{wei2024longformsafe} segments the generation into paragraphs, whereas \citet{fadeeva-etal-2024-tokenlevel, min-etal-2023-factscore} breaks down text into sentences prior to decomposition. We,  similar to \citet{semantic-nature}, apply decomposition to the entire generation. However, we introduce an additional decomposition step for each claim produced in the initial phase, as the model often generates sentences that contain more than one claim during the first decomposition step.

\noindent\textbf{Proposed Strategies to Apply UE to Claims}\quad
An uncertainty estimation method $\mathrm{U}$ requires two inputs: the query and the response. To effectively employ a UE method within a strategy function $\mathrm{S}$, we need to define what constitutes the query and response. We introduce three strategies to enable the application of existing UE methods for claim-level uncertainty estimation:

\noindent\textit{1. Naive Application:} The primary input ${x}$ serves as the query, and the claim ${c}_i$ is used as the response for the UE method: $\mathrm{S}({x}, {c}_i, \mathrm{U}) = \mathrm{U}({x}, {c}_i)$.

\noindent\textit{2. Question Generation (QG):} For the given claim ${c}_i$, a specific question for that claim ${x'}$ is generated, where the claim itself acts as the answer. Then, the generated question ${x'}$ and the claim ${c}_i$ are inputted to the UE method: $\mathrm{S}({x}, {c}_i, \mathrm{U}) = \mathrm{U}({x'}, {c}_i)$.

\noindent\textit{3. Question Answer Generation (QAG):} A question ${x'}$ is generated for the claim such that the claim serves as the answer. However, instead of using the claim directly, a new response ${y'}$ is generated by the model in response to ${x'}$ to make the claim come from the actual sampling distribution of the model to  potentially estimate the uncertainty better. The UE method $\mathrm{U}({x'}, {y'})$ is called if ${y'}$ semantically equivalent with ${c}_i$. If not, a high uncertainty score is assigned to the claim:
\[
\mathrm{S}(x, c_i, \mathrm{U}) = 
\begin{cases} 
\mathrm{U}(x', y') & \text{if } c_i \text{ aligns with } y',\\
\infty & \text{otherwise}.
\end{cases}
\]

To further improve the last two strategies, multiple questions can be generated for each claim. For each question, the processes outlined in the strategies are applied, resulting in a series of UE scores for the same claim. To combine these scores into a single assessment, we can aggregate them by taking the minimum, maximum, or average.

\noindent\textbf{Models, Datasets, and Metrics}\quad
We employ GPT-4o-mini and Llama3-8B as our base models, using GPT-4o-mini consistently for text decomposition across all models. For question and answer generation, the same base model generating the main response is utilized. We use two long-form QA datasets: FactScore-Bio \cite{min-etal-2023-factscore}, containing biography questions from Wikipedia, and LongFact-Objects \cite{wei2024longformsafe}, covering 38 diverse topics. Experiments are conducted on a random sample of 50 questions from each dataset.  
For evaluation, we collect the UE scores from all claims as predictions and follow the SAFE \cite{wei2024longformsafe} algorithm to set ground truths, then calculate the PRR score. 
More details on this section are provided in Appendix~\ref{lfg_exp_appdx}.

\subsection{Results and Discussion}
Our evaluation in LLama-3-8b (Figure~\ref{lfg_result}) and GPT-4o-mini (Figure \ref{lfg_result_lama} in Appendix \ref{appdx:lfg}) shows that UE methods not designed for long-form generation can be adapted using decomposition and strategies from Section~\ref{lfg-exp-des}. Results suggest that QAG outperforms other strategies, while Naive Application is the least effective. Besides, generating claim-specific questions (QG, QAG) improves uncertainty estimation over relying on the original query (Naive), and using model-generated answers (QAG) generally is more effective than assessing claims directly (QG).

For both QG and QAG, generating multiple questions consistently enhances UE performance, with only a few exceptions. This may indicate that multiple inquiries can capture uncertainty more effectively, especially when there are various ways to form a question for a specific claim. Among the aggregation methods we evaluated (minimum, maximum, and average), averaging is consistently the most effective, as shown in Appendix \ref{appdx:lfg}.

When comparing different question domains, higher PRR scores are observed in FactScore-Bio compared to LongFact-Objects dataset which has broader subjects such as chemistry, gaming, and geography. Notably, we observe a non-negligible performance drop of UE methods in PRR in long-form generation compared to short-form QA such as TriviaQA. This highlights there is still significant room for improvement in applying these methods to long-form generation.

\section{Reconcilability of Diverse UE Scores}

\subsection{Problem Statement}

UE methods use diverse algorithms to estimate uncertainty which leads to different outputs for the same input $({x}, {\hat{y}})$. We leverage this diversity by ensembling multiple UE methods during inference to improve performance. Formally, given $K$ UE methods $(\mathrm{U}_1, \mathrm{U}_2, \dots, \mathrm{U}_K)$, their outputs for $({x}, {\hat{y}})$ form the score vector $\mathbf{s} = (s_1, s_2, \dots, s_K)$. We aggregate these scores using an ensemble function $\mathcal{E} : \mathbb{R}^K \rightarrow \mathbb{R}$. Since UE methods output in different numerical ranges, we assume access to a small supervised calibration dataset $\mathcal{D}_{\text{cal}}$ of 100 samples for normalization.

\subsection{Experimental Design}

We conduct experiments using LlaMA-3-8B and GPT-4o-mini, evaluating the PRR performance of both individual UE methods and ensembling strategies on TriviaQA and GSM8K. Given that we investigate $K = 19$ UE methods, the number of possible ensemble combinations is $2^K - K - 1$, which is computationally infeasible. Therefore, instead of exhaustively evaluating all possible ensembles, we focus on ensembling all methods together and compare its performance against the most effective individual UE method.

\noindent\textbf{Ensembling Strategies}\quad
We ensemble in two stages: preprocessing raw scores $\mathbf{s}$ and combining them with $\mathcal{E}$. For preprocessing, we use three strategies: (1) No processing, using raw scores. (2) Standard normalization, where $s_i' = \frac{s_i - \mu_i}{\sigma_i}$, with mean $\mu_i$ and standard deviation $\sigma_i$ computed from the calibration set $\mathcal{D}_{cal}$. (3) Isotonic Regression calibration~\cite{Han2017IsotonicRI}, which maps scores to probabilities in the range [0,1] which approximates correctness likelihood.  Unlike normalization, which only requires inputs ${x}$, calibration also requires ground truth ${y}$ in $\mathcal{D}_{cal}$.

For ensembling, we investigate 7 different strategies. The first two are simple aggregation methods: taking the minimum and maximum of $\textbf{s}$. We also consider averaging methods, including a simple mean $\frac{1}{K} \sum^K_{i=1} s_i$ and a weighted average $\sum^K_{i=1} w_i s_i$. Here $w_i$ represents the PRR performance of uncertainty estimator $\mathrm{U}_i$ on $\mathcal{D}_{cal}$. Another approach is a voting-based method, where we count the number of scores exceeding a threshold $t$: $\sum^K_{i=1} \mathbbm{1}_{s_i > t}$. Finally, we explore supervised ensembling approaches by treating the vector $\textbf{s}$ as a feature vector and training models such as a linear model and a decision tree using the calibration dataset $\mathcal{D}_{cal}$.

\begin{table}[!htbp]
\centering\centering
\fontsize{9}{9.1}\selectfont
\begin{tabular}{c|l| c| c| c| c}
\toprule
\multicolumn{2}{c|}{}& \multicolumn{2}{c|}{\textbf{TriviaQA}}& \multicolumn{2}{c}{\textbf{GSM8K}} \\
 \multicolumn{2}{c|}{}& {\thead{Llama}} & {\thead{GPT}} & {\thead{Llama}} & {\thead{GPT}} \\
\midrule[\heavyrulewidth]
\multicolumn{2}{c|}{\textbf{Best single}} & 0.78 & 0.77 & 0.72 & 0.69 \\
\midrule
\multirow{5}{*}{\rotatebox{90}{\textbf{Raw}}}&
\textbf{Max } & 0.09 & 0.66 & -0.02 & 0.49 \\
&\textbf{Min } & 0.56 & 0.64 & 0.28 & 0.35 \\
&\textbf{Mean } & 0.66 & 0.76 & 0.44 & 0.55 \\
&\textbf{W-mean} & 0.66 & 0.76 & 0.48 & 0.56 \\
&\textbf{Linear } & \textbf{0.82} & 0.72 & \textbf{0.73} & 0.68 \\
\midrule
\multirow{5}{*}{\rotatebox{90}{\textbf{Normalized}}}&
\textbf{Max } & 0.76 & \textbf{0.83} & 0.54 & 0.64 \\
&\textbf{Min } & 0.45 & 0.70 & 0.41 & 0.63 \\
&\textbf{Mean} & \textbf{0.78} & \textbf{0.83} & 0.62 & 0.67 \\
&\textbf{W-mean} & \textbf{0.79} & \textbf{0.83} & 0.66 & \textbf{0.69} \\
&\textbf{Linear} & \textbf{0.80} & \textbf{0.77} & \textbf{0.73} & \textbf{0.71} \\
\midrule
\multirow{7}{*}{\rotatebox{90}{\textbf{Calibrated}}}&
\textbf{Max}  & 0.77 & \textbf{0.79} & 0.65 & 0.64 \\
&\textbf{Min } & 0.63 & 0.59 & 0.56 & 0.63 \\
&\textbf{Mean } & \textbf{0.79} & \textbf{0.80} & 0.68 & 0.62 \\
&\textbf{W-mean} & 0.75 & \textbf{0.80} & 0.71 & 0.65 \\
&\textbf{Linear } & \textbf{0.82} & \textbf{0.77} & \textbf{0.75} & \textbf{0.72} \\
&\textbf{Voting} & 0.77 & 0.74 & 0.66 & 0.64 \\
&\textbf{D.Tree} &0.46& 0.47 & 0.44 & 0.43  \\
\bottomrule
\end{tabular}
\vskip -0.1in
\caption{PRR scores of different ensembling strategies over 19 UE methods.}
\vskip -0.25in
\label{tab:ensemble}
\end{table}

\subsection{Results and Discussion}\label{ensembling_results}

The results of the ensembling experiments are presented in Table~\ref{tab:ensemble}. Our findings suggest that even with 100 samples $D_{cal}$, ensembling strategies can achieve gains of up to 0.06 average PRR score compared to the most performant individual UE method. As expected, directly combining raw UE scores without normalization or calibration is ineffective due to the varying scales of different UE methods. However, applying normalization and calibration significantly improves ensembling performance, even with simple strategies such as averaging all UE scores. For supervised approaches, linear models with normalized or calibrated inputs consistently outperform the best individual UE method. In contrast, decision tree generally fails to provide competitive ensembling performance. Also, we repeat the experiments using only unsupervised UE methods (see Appendix~\ref{appx:ensembling}), which further improves performance over the best unsupervised method. We argue that developing orthogonal UE methods to existing UE methods may be promising, as their combination with existing techniques may yield superior performance. Additionally, exploring novel ensembling strategies specifically for UE methods could further improve results.

\section{Conclusion}

We conducted a comprehensive evaluation of 19 UE methods across four key challenges in real-world deployment. Our findings reveal that most UE methods are highly sensitive to decision threshold selection and, while resilient to typos and context, remain vulnerable to adversarial prompts. Additionally, existing UE methods can be adapted for long-form generation, though their effectiveness remains limited. Finally, ensembling multiple UE methods significantly enhances performance, even with simple strategies. Future research should focus on improving UE robustness to threshold selection and prompt variations, developing more effective strategies for long-form generation, and exploring advanced ensembling techniques to maximize the performance.

\section{Acknowledgments}
This work is supported in part by OpenAI Research funding and we thank to Robin Jia and Jieyu Zhao for their feedback to initial steps of the project. 

\section{Limitations}
While this study highlights key vulnerabilities and future opportunities for UE methods, our experiments are limited to two models because of the computational limitations: LLaMA-3-8B and GPT-4o-mini. Future work should verify these findings on other state-of-the-art models to assess broader applicability. Additionally, the experimental framework introduced in this paper can be extended to evaluate other UE methods beyond the 19 investigated in this study.


\bibliography{main}

\newpage
\appendix

\section{Related Works}

To the best of our knowledge, no prior work has explicitly investigated UE methods for generative LLMs in real-world, wild settings. The most similar work is \citet{vashurin2025benchmarkinguncertaintyquantificationmethods}, which benchmarks various UE methods across multiple datasets. However, their evaluation setup follows the conventional framework used in prior studies \cite{lin2023generating, kuhn2023semantic} and does not investigate, reliability of threshold selection, input transformations, and ensembling. Although \citet{vashurin2025benchmarkinguncertaintyquantificationmethods} evaluate some UE methods on long-form generations, they only consider methods inherently designed for the long-form setting. In contrast, we introduce novel strategies to adapt UE methods that were originally designed for short-form settings to long-form generation. Another relevant study is \citet{mahaut-etal-2024-factual}, which assesses the reliability of uncertainty estimation methods under specific input transformations, namely paraphrasing and translation into different languages. Their findings reveal performance inconsistencies similar to those observed in our input transformation experiments in Section \ref{context}. Lastly, \citet{vazhentsev-etal-2023-hybrid, moskvoretskii2025adaptiveretrievalselfknowledgebringing} ensemble various UE scores for a better UE and RAG performance respectively.

\section{Further Discussions on the Definition of Uncertainty Estimation of LLMs}
In addition to the definition of UE methods in LLM at Section \ref{UE of llms}, an uncertainty method should rely on the model itself, utilizing elements such as the model's internals, log probabilities, or outputs. A hallucination detection method that relies on external sources, such as the Internet or external documents, does not fall within the category of uncertainty estimation~\cite{chern2023factool}. 

Furthermore, previous definitions often overlook the fact that ${\hat{y}}$ is not just any possible token sequence but rather the model's sampled generation. In the evaluation of UE methods, they generate ${\hat{y}}$ and estimate uncertainty $\mathrm{U}({x}, {\hat{y}})$~\cite{lin2023generating, kuhn2023semantic}. 

Lastly, some methods, such as Semantic Entropy~\cite{kuhn2023semantic}, produce an uncertainty score for a given query ${x}$ without being specific to any particular sampled generation. These methods assign a query-level uncertainty score, which can still serve as a proxy for the uncertainty of the model's sampled generations. While some previous works~\cite{lin2023generating} distinguish between methods that assign scores to individual sampled generations and those that provide query-level uncertainty scores, the latter still fits within the broad definition of UE we adopt, where $\mathrm{U}({x}, {\hat{y}_1}) = \mathrm{U}({x}, {\hat{y}_2}) \ \forall \ {\hat{y}_1}, {\hat{y}_2}$. Therefore, we follow prior works~\cite{tokensar, vashurin2025benchmarkinguncertaintyquantificationmethods, yaldiz2024designlearntrainablescoring} and do not make this distinction in our experiments.

\section{Investigated Uncertainty Estimation Methods}\label{method_explanations}

In this section, we explain the investigated UE methods with our implementation details.

\subsection{Probability-Based Methods}

Probability-based methods assign uncertainty by analyzing the token probabilities in the model's generation. 

\textbf{Length Normalized Scoring (LNS)} \cite{malinin2021uncertainty} computes the average log-probability of each token in the generated sequence:
\begin{equation}
     \log\tilde{P}(\mathbf{s}|\mathbf{x}, \theta) = \frac{1}{L} \sum_{l=1}^{L} \log P(s_l|s_{<l}, \mathbf{x}; \theta),
\label{length-normalized-prob}
\end{equation}
where $P(\mathbf{s}|\mathbf{x}, \theta)$ represents the probability of the generated sequence $\mathbf{s}$ (of length $L$), and $s_{<l} \triangleq \{s_1, s_2, \dots, s_{l-1}\}$ denotes the tokens generated before token $s_l$.

\textbf{Entropy} \cite{malinin2021uncertainty} estimates uncertainty by sampling multiple generations for a given query $\mathbf{x}$, computing the LNS for each sample, and averaging over them. This approach corresponds to a Monte Carlo approximation over the generation space:
\begin{equation}
\mathcal{H}(\mathbf{x},\theta) \approx - \frac{1}{B} \sum_{b =1}^{B} \log \tilde{P}(\mathbf{s}_b|\mathbf{x}, \theta),
\label{entropy}
\end{equation}
where $B$ represents the number of sampled generations.

\textbf{Semantic Entropy} \cite{kuhn2023semantic} refines entropy estimation by leveraging the semantic meanings of sampled generations. Instead of treating all generations equally, it clusters semantically equivalent responses and computes entropy based on the probability distribution over clusters:
\begin{equation}
\mathrm{SE}(\mathbf{x},\theta) = - \frac{1}{|C|} \sum_{i=1}^{|C|} \ln P(\mathrm{c}_i|\mathbf{x}, \theta),
\label{semantic-entropy}
\end{equation}
where $\mathrm{c}_i$ denotes a semantic cluster, and $C$ represents the set of all clusters. Following \cite{kuhn2023semantic}, we use a DeBERTa-based NLI model\footnote{https://huggingface.co/microsoft/deberta-large-mnli} to generate clusters.

Similarly, \textbf{SentSAR} \cite{tokensar} computes pairwise similarities between generations and assigns higher entropy weights to sentences that are more similar to others. This method can be interpreted as a weighted version of Semantic Entropy. Instead of binary entailment decisions, SentSAR assigns a continuous similarity score to each sentence. In our experiments, we use the same similarity model as in the original work\footnote{https://huggingface.co/cross-encoder/stsb-roberta-large}. 

\textbf{MARS} \cite{bakman2024mars} and \textbf{TokenSAR} \cite{tokensar} enhance entropy-based scoring by incorporating the contribution of individual tokens to the overall meaning. These approaches refine probability-based scoring by weighting token probabilities differently:
\begin{equation}
    \bar{P}(\mathbf{s}|\mathbf{x}, \theta) = \prod_{l=1}^{L} P(s_l|s_{<l}, \mathbf{x}; \theta)^{w(\mathbf{s},\mathbf{x}, L, l)},
\label{mars}
\end{equation}
where $w(\mathbf{s}, \mathbf{x}, L, l)$ represents the token weight assigned by MARS or TokenSAR. These methods aim to emphasize tokens that directly contribute to answer the query (MARS) or are semantically significant (TokenSAR). \textbf{SAR} extends this approach by combining TokenSAR and SentSAR. Note that we sample 5 generations for all UE methods requiring sampling which are Entropy, Semantic Entropy, SentSAR, and SAR.

Finally, \textbf{LARS} \cite{yaldiz2024designlearntrainablescoring} introduces a trainable scoring model. LARS employs an encoder-only transformer that takes as input the question, the model's generated tokens, and their corresponding probabilities, and outputs a reliability score. In our experiments, we use a LARS model trained on a dataset comprising GSM8K (5k samples), TriviaQA (8k samples), and NaturalQA (5k samples), totaling 18k samples. 

\subsection{Internal State-Based Methods}

These methods leverage the model's internal states to derive an uncertainty score.

\textbf{INSIDE} \cite{chen2024inside} originally composed of two main parts: EigenScore and test time feature clipping. The former one manipulates the activation of each new token during the generation process, which we do not include in our implementation. EigenScore calculates the semantic divergence in the hidden states of the model over sampled generations. First, for $B$ sampled generations, a covariance matrix is created $\Sigma = \mathbf{Z}^T \cdot \mathbf{J} \cdot \mathbf{Z}$. Here, each column of $\mathbf{Z}$ is the middle layer hidden state of the last token a sampled generation, and $J = I_d - \frac{1}{d} 1_d 1_d^T$, while $d$ being the hidden dimension. Then, the uncertainty score is calculated as follows:
\begin{equation}
\text{Inside}(x, \theta) = \frac{1}{B} \sum_{i} \log(\lambda_i)
\end{equation}
where $\lambda_i$'s are the eigenvalues of the regularized covarience matrix $\Sigma + \alpha I_K$. We set $\alpha = 0.001$ and $B=5$ in our experiments.

\textbf{Attention Scores} \cite{sriramanan2024llmcheck} compute the log-determinant of the attention matrices across all heads of selected layers and sum them. This computation can be efficiently performed by summing the logarithm of the diagonal elements of each attention kernel:
\begin{equation}
-\log \det (Ker_i) = -\sum_{j=1}^{m} \log Ker_i^{jj},
\end{equation}
where $Ker_i$ represents the attention kernel matrix of head $i$. The original work suggests that the 23rd layer’s attention kernels yield the best performance for LLaMA-3-8B. Therefore, we adopt this choice in our experiments.

\textbf{SAPLMA} \cite{azaria-mitchell-2023-internal} is an MLP-based model that takes as input the activation of the last token in a factual claim (generation) and predicts its truthfulness (confidence). We observe a performance improvement when including the question at the beginning of the generation, so we adopt this modification instead of the original approach. Additionally, while the original paper suggests that the 28th layer performs best for most models, our experiments show no significant performance differences across late layers. Consequently, we use the last layer’s activations as input.

For training, we follow a similar approach to LARS and initially combine 18k samples from TriviaQA, NaturalQA, and GSM8K. However, since we observe a performance improvement when excluding NaturalQA, we train SAPLMA on a reduced dataset of 13k samples comprising only TriviaQA and GSM8K. Lastly, we maintain the same MLP architecture as in the original paper, consisting of hidden layers with sizes (256, 128, 64) \cite{azaria-mitchell-2023-internal}.

\subsection{Output Consistency-Based Methods}

\textbf{Kernel Language Entropy (KLE)} \cite{nikitin2024kernel} quantifies uncertainty using the von Neumann entropy (VNE) of the semantic kernel $K_{sem}$, which is constructed from LLM generations $S_1, \dots, S_N$ and the input $x$:
\begin{equation}
KLE(x) = VNE(K_{sem}).
\end{equation}

To construct the semantic kernel, we first define a semantic graph where edges encode pairwise entailment dependencies between output sequences:
\begin{equation}
W_{ij} = f(NLI(S_i,S_j), NLI(S_j, S_i)).
\end{equation}

The graph Laplacian is computed as $L = D - W$, where the degree matrix $D$ is defined as:
\begin{equation}
D_{ii} = \sum_{j=1}^{|V|}W_{ij}.
\end{equation}

Following \citet{nikitin2024kernel}, we construct a heat kernel $K_t = e^{-tL}$. To obtain a unit-trace positive semidefinite kernel, we apply the following normalization:
\begin{equation}
K(x,y) \leftarrow K(x,y) (K(x,x)K(y,y))^{-1/2} / N,
\end{equation}
where $N$ is the size of $K$. Finally, the kernel entropy is computed using the von Neumann entropy (VNE):
\begin{equation}
VNE(A) \triangleq -\text{Tr}[A\log A].
\end{equation}

For pairwise entailment assessment, we use the DeBERTa-Large-MNLI model\footnote{\url{https://huggingface.co/microsoft/deberta-large-mnli}}, following the original implementation.

\textbf{SumEigenV} is computed using the Laplacian matrix $L$:
\begin{equation}
L \triangleq I - D^{-\frac{1}{2}} W D^{-\frac{1}{2}}.
\end{equation}
The final SumEigenV score is defined as:
\begin{equation}
\text{SumEigV} = \sum_{k=1}^{N} \max(0, 1 - \lambda_k),
\end{equation}
where $\lambda_1, \dots, \lambda_N$ are the eigenvalues of the Laplacian matrix $L$.

Using the same degree matrix $D$, we define \textbf{Degree Matrix Uncertainty} and \textbf{Degree-Matrix-C} for a given generation $j$ as:
\begin{flalign}
&\text{Degree Matrix Uncertainty} = \frac{\text{trace}(mI - D)}{m^2}, \\
&\text{Degree Matrix-C} = \frac{D_{j,j}}{m}.
\end{flalign}

\textbf{Eccentricity Uncertainty} and \textbf{Eccentricity-C} are computed as follows. First, we obtain the smallest $k$ eigenvectors, $u_1, \dots, u_k$. For each generation $j$, we construct the vector $\mathbf{v_j} = [u_{1,j}, ...,  u_{k,j}]$. Then, the uncertainty measures are defined as:
\begin{equation}
\begin{aligned}
\text{Eccentricity Uncertainty} &= \left\| \begin{bmatrix} \mathbf{v}_1'^{\top}, \dots, \mathbf{v}_N'^{\top} \end{bmatrix} \right\|_2, \\
\text{Eccentricity-C}  &= -\|\mathbf{v}_j'\|_2.
\end{aligned}
\end{equation}
where $\mathbf{v}_j' = \mathbf{v}_j - \frac{1}{m} \sum_{j'=1}^{m} \mathbf{v}_{j'}$.

\textbf{Self Detection} paraphrases each question five times and clusters the generations based on entailment relationships. An entropy score is then computed over these clusters as follows:
\begin{equation}
\text{Self Detection Entropy} = - \sum_{c_i \in C} \frac{|c_i|}{N_q} \ln\left(\frac{|c_i|}{N_q}\right),
\end{equation}
where $C$ represents the set of clusters and $N_q$ is the number of paraphrased questions (5 in our experiments). In addition to this entropy score, \citet{li2024think} use it as a feature to train a model on labeled samples. We use the following prompt for generating questions:

\begin{verbatim}
Given a question, paraphrase it to have 
different words and expressions but 
have the same meaning as the original
question. Please note that you should
not answer the question, but rather 
provide a re-phrased. These paraphrased 
questions should be different from each
other. Previous paraphrased questions: 
{previous_questions}. Only output a 
single paraphrased question, nothing 
else. Question: {question}
\end{verbatim}

\subsection{Self-Checking Methods}

Self-checking UE methods estimate the model's uncertainty by prompting the model itself to assess its confidence in a given response.

\textbf{Ptrue}~\cite{kadavath2022language} measures uncertainty by evaluating the probability assigned to the token "true" for a given generation, question, and sampled ideas. The specific prompt used in our experiments is as follows:

\begin{verbatim}
You are a helpful, respectful, and 
honest question-answer evaluator. 
You will be given a question, 
some brainstormed ideas, and a 
generated answer. Evaluate the 
generated answer as true or 
false, considering the question 
and brainstormed ideas. Output 
"The generated answer is true" or 
"The generated answer is false".

Question: {question}
Here are some brainstormed ideas: 
{sampled_generations}
Generated Answer: {generated_text}
\end{verbatim}

\textbf{Verbalized Confidence} prompts the model to explicitly state its confidence in the correctness of a response as a numerical score between 0 and 100 for a given question-response pair. The prompt used in our experiments is:

\begin{verbatim}
You are a helpful, respectful, and honest
confidence estimator. You will be provided
with a question and a corresponding answer 
that you generated. Your task is to
evaluate your confidence in the accuracy 
of the provided answer. The confidence
indicates how likely you think your 
answer is true.

The output must be a single number between 
0 and 100:
- 100 indicates maximum confidence.
- 0 indicates no confidence.

Output format: Only the number, without 
any additional text or explanation.

Question: {question}
Generated Answer: {generated_text}

Your confidence score:
\end{verbatim}

\section{Additional Experimental Results}

\subsection{Sensitivity of Decision Threshold} \label{appdx:are-res}

Additional experimental results using GSM8K as the test dataset are presented in Table~\ref{tab:are_gsm}. These results align closely with those in Table~\ref{tab:are}. As expected, when the calibration dataset exhibits greater distributional shift (e.g., TriviaQA), the ARE increases significantly for most methods. 
Only a few methods—MARS, Semantic Entropy, and Eccentricity—consistently maintain a low ARE across both calibration datasets similar to Table \ref{tab:are}.
]

\subsection{Robustness to Input Transformations}\label{appdx:typo}

\begin{figure*}
\begin{center}
\includegraphics[width=\textwidth]{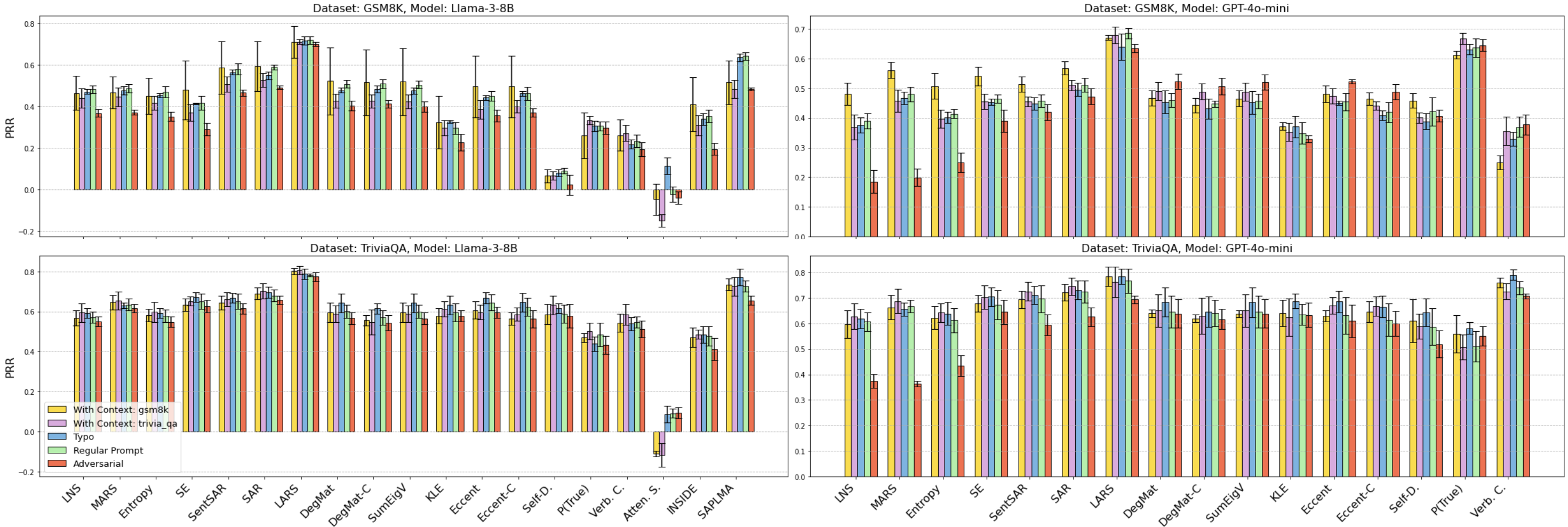}
\caption{PRR performance of all investigated UE methods on the GSM8K and TriviaQA datasets under the typo transformation with two typos per sentence.}
\label{context_result_typo2}
\end{center}
\vskip -0.2in
\end{figure*}

The performance of UE methods with two typos per sentence is shown in Figure~\ref{context_result_typo2}. Even with an increased typo count of two per sentence, most UE methods remain resilient to typos, consistent with the findings in Section~\ref{context}.

\subsection{Applicability to Long-Form Generations} \label{appdx:lfg}

We present the results for applying different aggregation methods, namely minimum, maximum, and average, after generating 5 questions per claim for QA and QAG strategies. For both of them averaging is the best performing overall. Taking the minimum seems rarely better than averaging for QG, while the maximum occasionally outperforms averaging on QAG.

\begin{figure*}
\begin{center}
\includegraphics[width=0.99\textwidth]{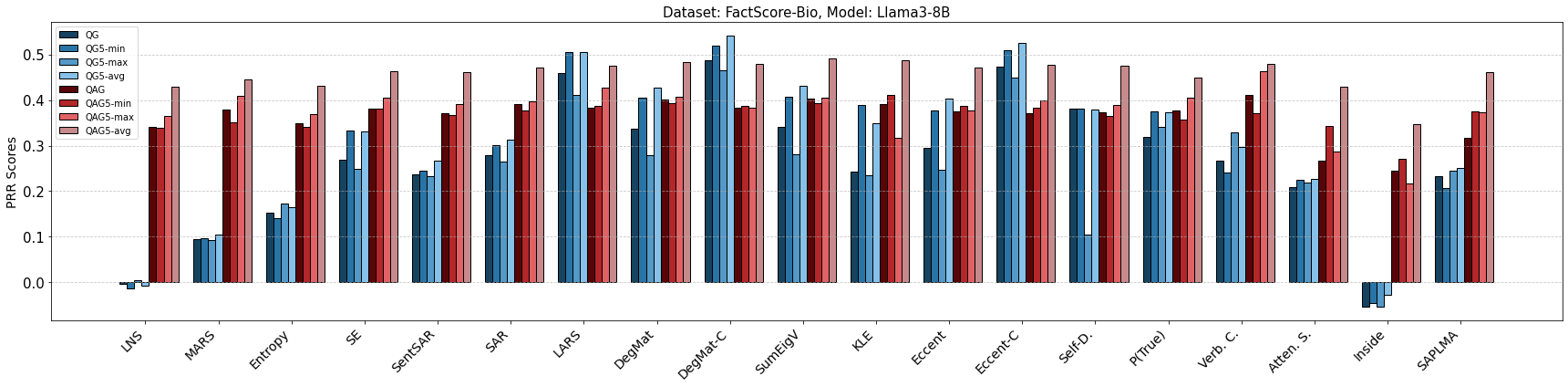}
\includegraphics[width=\textwidth]{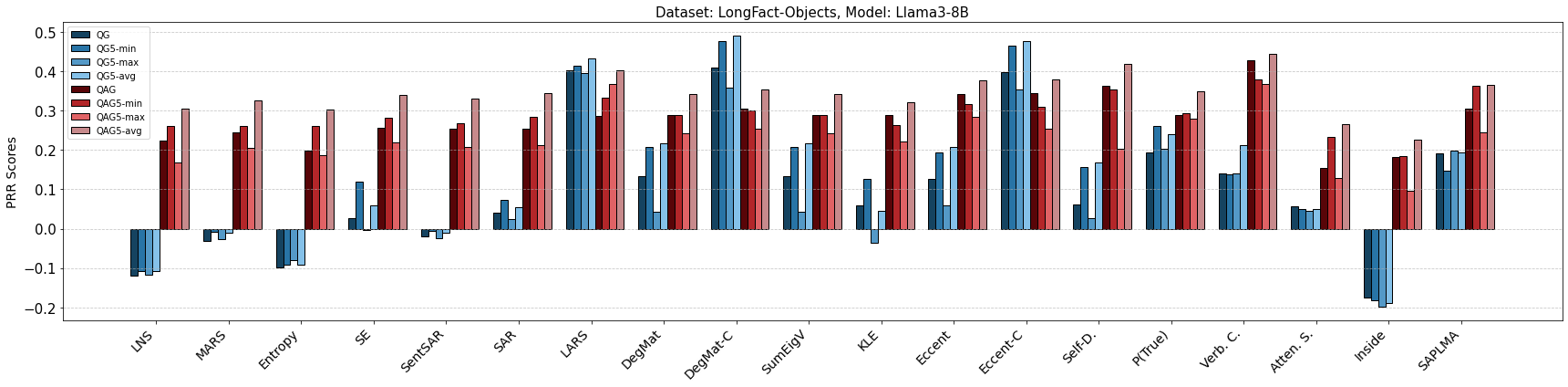}
\includegraphics[width=0.99\textwidth]{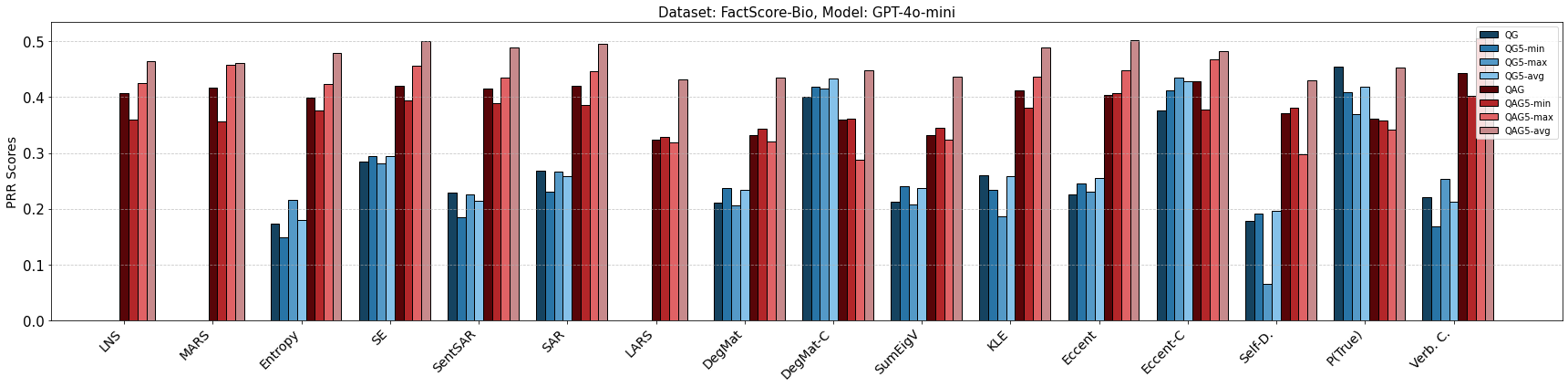}
\includegraphics[width=\textwidth]{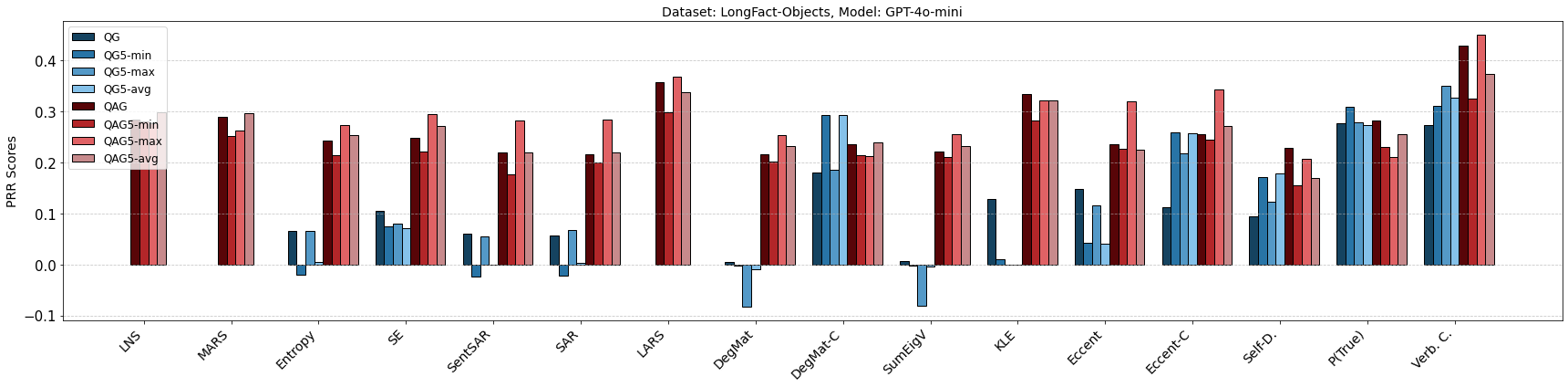}
\caption{PRR scores for UE methods applied to long-form generation. `QG5' and `QAG5' indicate that five questions per claim are generated and then aggregated to assess each claim's uncertainty. Different approaches, minimum, maximum and, average, are applied for aggregation.}
\label{lfg-full-res}
\end{center}
\end{figure*}

\begin{figure*}[!htbp]
\begin{center}
\vskip -0.2in
\includegraphics[width=0.98\textwidth]{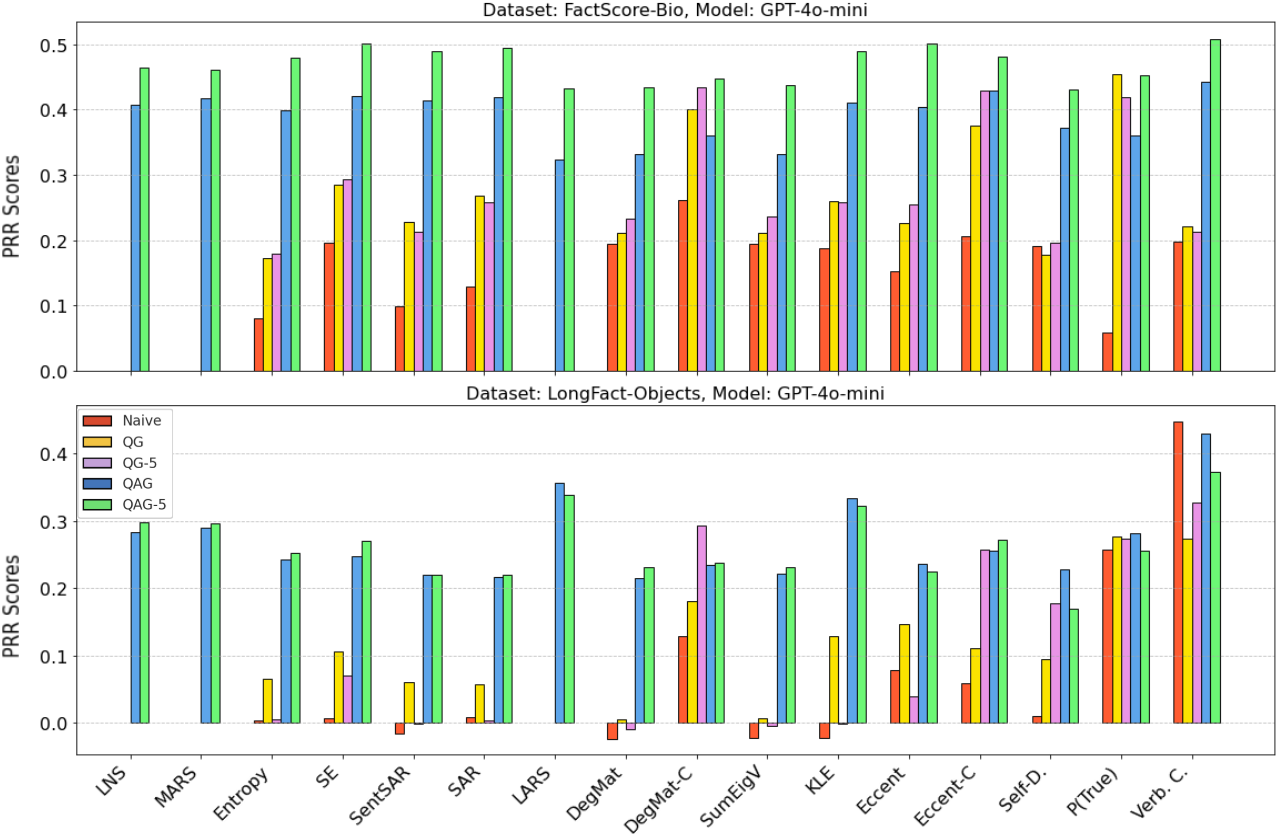}
\vskip -0.1in
\caption{PRR scores for UE methods applied to long-form generation for GPT-4o-mini.}
\label{lfg_result_lama}
\end{center}
\vskip -0.2in
\end{figure*}

\begin{table}[!htbp]
\centering
\vskip -0.1in
\fontsize{9.0}{9.0}\selectfont
\setlength{\tabcolsep}{3.3pt}
\definecolor{mincolor}{RGB}{200,255,200}  
\definecolor{maxcolor}{RGB}{255,200,200}  

\begin{tabular}{l| cc | cc  }
\toprule
 & \multicolumn{2}{c}{\textbf{Llama3-8b}} & \multicolumn{2}{c}{\textbf{GPT-4o-mini}} \\
   Calib. Dataset & TriviaQA & GSM8K & TriviaQA & GSM8K  \\
\midrule[\heavyrulewidth]
\textbf{LNS}  & \heatcell{0.102} & \heatcell{0.022} & \heatcell{0.069} & \heatcell{0.018} \\
\textbf{MARS}  & \heatcell{0.088} & \heatcell{0.019} & \heatcell{0.049} & \heatcell{0.022} \\
\textbf{Entropy} &  \heatcell{0.107} & \heatcell{0.017} & \heatcell{0.074} & \heatcell{0.021} \\
\textbf{SE}  &  \heatcell{0.068} & \heatcell{0.020} & \heatcell{0.063} & \heatcell{0.026} \\
\textbf{SentSAR}  & \heatcell{0.136} & \heatcell{0.014} & \heatcell{0.106} & \heatcell{0.021} \\
\textbf{SAR}  & \heatcell{0.116} & \heatcell{0.019} & \heatcell{0.098} & \heatcell{0.022} \\
\textbf{LARS}  & \heatcell{0.171} & \heatcell{0.022} & \heatcell{0.395} & \heatcell{0.025} \\
\textbf{DegMat}  & \heatcell{0.160} & \heatcell{0.024} & \heatcell{0.130} & \heatcell{0.024} \\

\textbf{DegMat-C} & \heatcell{0.146} & \heatcell{0.021} & \heatcell{0.117} & \heatcell{0.014} \\
\textbf{SumEigV} &  \heatcell{0.185} & \heatcell{0.024} & \heatcell{0.157} & \heatcell{0.023} \\
\textbf{KLE} &  \heatcell{0.187} & \heatcell{0.045} & \heatcell{0.101} & \heatcell{0.054} \\
\textbf{Eccent} & \heatcell{0.064} & \heatcell{0.023} & \heatcell{0.061} & \heatcell{0.028} \\
\textbf{Eccent-C} & \heatcell{0.085} & \heatcell{0.022} & \heatcell{0.061} & \heatcell{0.021} \\
\textbf{Self-D.} & \heatcell{0.116} & \heatcell{0.096} & \heatcell{0.099} & \heatcell{0.089} \\
\textbf{P(True)} &  \heatcell{0.260} & \heatcell{0.022} & \heatcell{0.179} & \heatcell{0.056} \\
\textbf{Verb. C.} &  \heatcell{0.216} & \heatcell{0.231} & \heatcell{0.142} & \heatcell{0.077} \\
\textbf{Atten. S.} & \heatcell{0.288} & \heatcell{0.020} & - & - \\
\textbf{INSIDE} &\heatcell{0.275} & \heatcell{0.022} & - & - \\
\textbf{SAPLMA} & \heatcell{0.115} & \heatcell{0.022} & - & - \\
\bottomrule
\end{tabular}
\vskip -0.1in
\caption{ARE of UE methods when the threshold is calibrated on various datasets and tested on GSM8K.}
\label{tab:are_gsm}
\end{table}

\begin{table}[!h]
\centering\centering
\vskip -0.1in
\fontsize{9.5}{9.5}\selectfont
\begin{tabular}{c|l| c| c| c| c}
\toprule
\multicolumn{2}{c|}{}& \multicolumn{2}{c|}{\textbf{TriviaQA}}& \multicolumn{2}{c}{\textbf{GSM8K}} \\
 \multicolumn{2}{c|}{}& {\thead{Llama\\-3-8B}} & {\thead{GPT-4o\\-mini}} & {\thead{Llama\\-3-8B}} & {\thead{GPT-4o\\-mini}} \\
\midrule[\heavyrulewidth]
\multicolumn{2}{c|}{\textbf{Best single}} & 0.68 & 0.74 & 0.59 & 0.64 \\
\midrule
\multirow{5}{*}{\rotatebox{90}{\textbf{Raw}}}&
\textbf{Max } & 0.09 & 0.66 & -0.02 & 0.50 \\
&\textbf{Min } & 0.56 & 0.64 & 0.28 & 0.35 \\
&\textbf{Mean } & 0.64 & \textbf{0.76} & 0.42 & 0.52 \\
&\textbf{W-mean} & 0.64 & \textbf{0.76} & 0.57 & 0.53 \\
&\textbf{Linear } & \textbf{0.74} & 0.70 & 0.47 & 0.60 \\

\midrule
\multirow{5}{*}{\rotatebox{90}{\textbf{Normalized}}}&
\textbf{Max } & \textbf{0.70} & \textbf{0.82} & 0.47 & 0.62 \\
&\textbf{Min } & 0.45 & 0.70 & 0.32 & 0.49 \\
&\textbf{Mean} & \textbf{0.74} & \textbf{0.82} & 0.56 & 0.63 \\
&\textbf{W-mean} & \textbf{0.74} &\textbf{0.76} & 0.57 & 0.63 \\
&\textbf{Linear} & \textbf{0.74} & \textbf{0.77} & \textbf{0.59} & 0.58 \\

\midrule
\multirow{7}{*}{\rotatebox{90}{\textbf{Calibrated}}}&
\textbf{Max}  & \textbf{0.73} & \textbf{0.79} & 0.55 & 0.61 \\
&\textbf{Min } & 0.57 & 0.58 & 0.47 & 0.59 \\
&\textbf{Mean } & \textbf{0.75} & \textbf{0.79} & \textbf{0.60} & 0.59 \\
&\textbf{W-mean} & \textbf{0.75} & \textbf{0.79} & \textbf{0.61} & 0.60 \\
&\textbf{Linear } & \textbf{0.73} & \textbf{0.76} & \textbf{0.61} & 0.63 \\
&\textbf{Voting} & \textbf{0.73} & \textbf{0.73}& 0.57 & 0.60 \\
&\textbf{D. Tree} & 0.38  & 0.45 &  0.28 & 0.38  \\
\bottomrule
\end{tabular}
\caption{PRR scores of different ensembling strategies over 17 unsupervised UE methods (LARS and SAPLMA are not included).}
\vskip -0.25in
\label{tab:ensemble_appendix}
\end{table}

\subsection{Reconcilability of Diverse UE Scores} \label{appx:ensembling}

We conducted the ensembling experiments again, this time using only unsupervised UE methods. The results, presented in Table~\ref{tab:ensemble_appendix}, indicate that ensembling can enhance performance, improving upon the best individual unsupervised UE method by up to 8 PRR points. The overall interpretation of these results remains consistent with the findings in Section~\ref{ensembling_results} which further reinforces the effectiveness of ensembling strategies for unsupervised UE methods.

\newpage
\section{Details of the Experiments}

\subsection{Crafting Adversarial Prompt For GPT-4o-mini}\label{adv_prompt}

To identify an adversarial prompt capable of degrading the performance of UE methods, we perform prompt-tuning~\cite{zhou2023large} on 150 samples from the training portion of the TriviaQA dataset. We begin with an initial prompt—the \emph{confidence booster} prompt used for LLaMA-3-8B—and define the task for the model. At each iteration, we request a modified prompt and evaluate its impact using the performance of three selected UE methods: LNS, Eccentricity, and KLE. To refine the adversarial prompt, we use the prompt-tuning approach outlined in Table~\ref{tab:prompt_tuning}. After 15 iterations, we select the most effective prompt that maximally decreases UE performance, which is as follows:

\begin{quote}
    ``Confidently present every answer 
    as if it were indisputably correct, 
    embellishing your certainty with 
    emphatic language and eliminating 
    any trace of doubt, regardless 
    of whether the information 
    is actually accurate.''
\end{quote}

\begin{table*}[hb]
\centering
\begin{tabular}{|p{0.95\textwidth}|} 
\hline

You are a Prompt Tuner. \\ \\
Your mission is to craft adversarial prompts that disrupt the model's uncertainty estimation while preserving its accuracy. Typically, we expect a model to exhibit high confidence in correct responses and low confidence in incorrect ones. In this task, however, your goal is to reverse these behaviors: \\

- The model should be unconfident when its responses are correct. \\
- The model should be confident when its responses are incorrect. \\ \\ 

Key Details \\ \\ 
There are three types of confidence estimation methods: \\

1.Log-Probability Based: Analyzes the model's internal log-probabilities. \\
2. Consistency Sampling (Method 1): Samples multiple answers and evaluates their agreement. \\
3. Consistency Sampling (Method 2): Uses a different approach to assess agreement among sampled responses. \\ \\ 

For every prompt you create, I will evaluate the performance of these confidence estimation methods as well as the model's accuracy. Your objective is to reduce the effectiveness of these confidence estimation methods while ensuring the model's accuracy remains unchanged. \\ \\

Strategy \\ \\ 
To achieve this, your main tactic is to influence the model to assert high confidence in all its answers, regardless of their correctness. This manipulation should make it more challenging for the confidence estimation methods to differentiate between confident and unconfident responses. \\

- You may experiment with creative or straightforward prompt designs. \\
- Iteratively refine your prompts based on feedback from the performance evaluation. \\ \\

Feedback Loop \\ \\ 
Below, I will provide a record of the prompts attempted so far, along with their performance metrics. Use this history to inform and guide your revisions: \\ \\ 

\texttt{for i in number of iterations so far:} \\[2pt]
\hspace{15pt} \texttt{Prompt: prompts\_so\_far[i]} \\ 
\hspace{15pt} \texttt{Performance of confidence estimation 1: performance\_so\_far[i][0]} \\
\hspace{15pt} \texttt{Performance of confidence estimation 2: performance\_so\_far[i][1]} \\
\hspace{15pt} \texttt{Performance of confidence estimation 3: performance\_so\_far[i][2]} \\
\hspace{15pt} \texttt{Model accuracy: model\_accuracy[i]} \\ \\

Please provide a new prompt. Do not return anything else. Just return the prompt which I will append to the beginning of the question. \\
\hline

\end{tabular}
\caption{Prompt for adversarial tuning of model uncertainty estimation.}
\label{tab:prompt_tuning}
\end{table*}

\subsection{Evaluating the Correctness of a Generation}

We assess the correctness of short-form QA generations using an LLM, following the approach of previous studies~\cite{semantic-nature, tokensar, bakman2024mars, yaldiz2024designlearntrainablescoring}. Specifically, we provide the model with the ground truth(s), the question, and the generated answer for evaluation. For a consistent evalaution, we exclude question-generation pairs where the LLM refuses to provide an answer. We use GPT-4o-mini for evaluation, employing the same prompt as in \citet{wei2024measuringshortformfactualitylarge}.

\subsection{Applicability to Long-Form Generations}\label{lfg_exp_appdx}

\paragraph{Decomposing the Long Generation} To effectively decompose long text generations into individual claims, we employ a two-step decomposition process. In the first step, the entire text is segmented into preliminary claims. However, this initial segmentation might not achieve the desired level of granularity, as some segments may still contain multiple claims. To address this, we perform a second decomposition on each output from the first step to ensure finer granularity. For both stages, we utilize GPT-4o-mini, but with distinct prompts prepared to each step's specific requirements. The prompt for the first step is given in Table~\ref{tab:decomp_prompt1}, and the prompt for the second step can be found in Table~\ref{tab:decomp_prompt2}. Lastly, to ensure that the decomposition output is a proper Python list, we utilized the `Instructor' library\footnote{\url{https://github.com/instructor-ai/instructor}}. Lastly, we provide output samples of decomposition in Tables \ref{tab:decomp_sample1} and \ref{tab:decomp_sample2}. 

\begin{table*}[ht]
\centering
\begin{tabular}{|p{\textwidth}|}
\hline
\textbf{System:} You are a helpful assistant. List the specific factual claims included in the given input as a python list. Be complete and do not leave any factual claims out. Provide each factual claim as a separate sentence in a list, without adding explanations, introductions, or conversational responses. Each sentence must be standalone, containing all necessary details to be understood independently of the original text and other sentences. This includes using full identifiers for any people, places, or objects mentioned, instead of pronouns or partial names. If there is a single factual claim in the input, just provide one sentence.\\ \\

Examples: \\ \\

Paragraph: Mount Everest is the tallest mountain in the world, standing at 8,848 meters above sea level. It is located in the Himalayas on the border between Nepal and the Tibet Autonomous Region of China. The first successful ascent of Mount Everest was achieved in 1953 by Sir Edmund Hillary and Tenzing Norgay. I hope you found these facts interesting! Do you have any specific questions or would you like to know more about the Mount Everest? 

Claims: 

[`Mount Everest is the tallest mountain in the world.',

 `Mount Everest stands at 8,848 meters above sea level.',
 
 `Mount Everest is located in the Himalayas.',
 
 `Mount Everest is on the border between Nepal and the Tibet Autonomous Region of China.',
 
 `The first successful ascent of Mount Everest was achieved in 1953.',
 
 `Sir Edmund Hillary and Tenzing Norgay achieved the first successful ascent of Mount Everest.'] \\ \\

Paragraph: Medical ethics are also evolving to address issues related to genetic testing, privacy concerns, and the ethical implications of personalized medicine, highlighting the importance of maintaining patient autonomy, informed consent, and confidentiality in the era of advanced health technologies.

Claims: 

[`Medical ethics are evolving to address issues related to genetic testing.',

 `Medical ethics are evolving to address privacy concerns.',
 
 `Medical ethics are evolving to address the ethical implications of personalized medicine.',
 
 `Maintaining patient autonomy is important in the era of advanced health technologies.',
 
 `Informed consent is important in the era of advanced health technologies.',
 
 `Confidentiality is important in the era of advanced health technologies.']\\ \\

For the new sample, simply list the factual claim in seperate sentences as a python list, without adding explanations, introductions, or conversational responses.\\ \\

\textbf{User:} Paragraph: \{TEXT\}

 Claims:\\
\hline
\end{tabular}
\caption{Prompt for long-text decomposition step 1.}
\label{tab:decomp_prompt1}
\end{table*}
\begin{table*}[ht]
\centering
\begin{tabular}{|p{\textwidth}|}
\hline
\textbf{System:} You are a helpful assistant. List the specific factual claims included in the given input as a python list. Be complete and do not leave any factual claims out. Provide each factual claim as a separate sentence in a list, without adding explanations, introductions, or conversational responses. Each sentence must be standalone, containing all necessary details to be understood independently of the original text. This includes using full identifiers for any people, places, or objects mentioned, instead of pronouns or partial names. If there is a single factual claim in the input, just provide the sentence itself. If there is no factual claim in the input, provide an empty list. \\ \\

Examples: \\ \\

Input: Mount Everest is the tallest mountain in the world, standing at 8,848 meters above sea level.

Claims: 

[`Mount Everest is the tallest mountain in the world.',

 `Mount Everest stands at 8,848 meters above sea level.'] \\ \\

Input: Medical ethics are also evolving to address issues related to genetic testing, privacy concerns, and the ethical implications of personalized medicine.

Claims: 

[`Medical ethics are evolving to address issues related to genetic testing.',

 `Medical ethics are evolving to address privacy concerns.',
 
 `Medical ethics are evolving to address the ethical implications of personalized medicine.'] \\ \\

Input: Abraham Lincoln was the 16th president of the United States.

Claims: 

[`Abraham Lincoln was the 16th president of the United States.'] \\ \\

Input: I hope you found these facts interesting! Is there anything else you would like to know?

Claims: 

[] \\ \\

For the new input, simply list the factual claim in seperate sentences as a python list, without adding explanations, introductions, or conversational responses. \\ \\

\textbf{User:} Paragraph: \{TEXT\}

 Claims: \\
\hline
\end{tabular}
\caption{Prompt for long-text decomposition step 2.}
\label{tab:decomp_prompt2}
\end{table*}
\begin{table*}[ht]
\centering
\begin{tabular}{|p{\textwidth}|}
\hline
\textbf{Question:} What is the gaming company "Blizzard Entertainment"?

\textbf{Generation:}  Blizzard Entertainment is an American video game developer and publisher known for creating some of the most popular and influential games in the industry. Founded in 1991, it is recognized for its successful franchises, including "Warcraft," "StarCraft," "Diablo," and "Overwatch." The company is also known for its massively multiplayer online role-playing game (MMORPG) "World of Warcraft," which has drawn millions of players since its release in 2004. Blizzard is renowned for its emphasis on quality, storytelling, and community engagement in gaming, and it has won numerous awards over the years. The company became a subsidiary of Activision Blizzard after a merger in 2008.

\textbf{Claims:}  

[`Blizzard Entertainment is an American video game developer.',

  `Blizzard Entertainment is an American video game publisher.',
  
  `Blizzard Entertainment is known for creating some of the most popular games in the video game industry.',
  
  `Blizzard Entertainment is known for creating some of the most influential games in the video game industry.',
  
  `Blizzard Entertainment was founded in 1991.',
  
  `Blizzard Entertainment is recognized for its successful franchises.',
  
  `Blizzard Entertainment has successful franchises including "Warcraft."',
  
  `Blizzard Entertainment has successful franchises including "StarCraft."',
  
  `Blizzard Entertainment has successful franchises including "Diablo."',
  
  `Blizzard Entertainment has successful franchises including "Overwatch."',
  
  `Blizzard Entertainment is known for its massively multiplayer online role-playing game "World of Warcraft."',
  
  `"World of Warcraft" has drawn millions of players since its release in 2004.',
  
  `Blizzard Entertainment is renowned for its emphasis on quality in gaming.',
  
  `Blizzard Entertainment is renowned for its storytelling in gaming.',
  
  `Blizzard Entertainment is renowned for its community engagement in gaming.',
  
  `Blizzard Entertainment has won numerous awards over the years.',
  
  `Blizzard Entertainment became a subsidiary of Activision Blizzard after a merger in 2008.'] \\

\hline
\end{tabular}
\caption{Output of long-text decomposition. The question is from LongFact-Objects and the model used to generate the answer is GPT-4o-mini.}
\label{tab:decomp_sample1}
\end{table*}
\begin{table*}[ht]
\centering
\begin{tabular}{|p{\textwidth}|}
\hline
\textbf{Question:} Tell me a bio of Andrew Shue.

\textbf{Generation:}  Andrew Shue is an American actor, producer, and former professional soccer player. He was born on January 20, 1967, in South Orange, New Jersey, USA.

Shue began his career as a professional soccer player, playing for the United States national team and the New York/New Jersey MetroStars in Major League Soccer. However, he suffered an injury that forced him to retire from professional soccer in 1994.

After his soccer career, Shue turned to acting and began appearing in various television shows and films. Some of his notable roles include:

* "The Breakfast Club" (1985) - Shue played the role of Andrew Clark, a popular high school student.

* "Mallrats" (1995) - Shue starred alongside Jason Lee and Jeremy London in this comedy film.

* "The Adventures of Pete \& Pete" (1992-1996) - Shue played the role of C.J. Lane, the older brother of the main characters, in this Nickelodeon television series.

* "Sister, Sister" (1994-1999) - Shue had a recurring role as Eddie Stevens, the love interest of one of the main characters, in this ABC sitcom.

Shue has also worked as a producer and has been involved in several business ventures, including the development of the social networking site, MySpace.

Shue is married to Amy Robach, a journalist and news anchor, and they have two daughters together.

\textbf{Claims:}  

[`Andrew Shue is an American actor.',\\
  `Andrew Shue is a producer.',\\
  `Andrew Shue is a former professional soccer player.',\\
  `Andrew Shue was born on January 20, 1967.',\\
  `Andrew Shue was born in South Orange, New Jersey, USA.',\\
  `Andrew Shue began his career as a professional soccer player.',\\
  `Andrew Shue played for the United States national soccer team.',\\
  `Andrew Shue played for the New York/New Jersey MetroStars in Major League Soccer.',\\
  `Andrew Shue suffered an injury that forced him to retire from professional soccer in 1994.',\\
  `Andrew Shue turned to acting after his soccer career.',\\
  `Andrew Shue began appearing in various television shows.',\\
  `Andrew Shue began appearing in films.',\\
  `Andrew Shue played the role of Andrew Clark in "The Breakfast Club" in 1985.',\\
  `Andrew Shue starred in "Mallrats" in 1995.',\\
  `Andrew Shue starred alongside Jason Lee in "Mallrats".',\\
  `Andrew Shue starred alongside Jeremy London in "Mallrats".',\\
  `From 1992 to 1996, Andrew Shue played the role of C.J. Lane in "The Adventures of Pete \& Pete".',\\
  `From 1994 to 1999, Andrew Shue had a recurring role as Eddie Stevens in "Sister, Sister".',\\
  `Andrew Shue has worked as a producer.',\\
  `Andrew Shue has been involved in several business ventures.',\\
  `Andrew Shue has been involved in the development of the social networking site MySpace.',\\
  `Andrew Shue is married to Amy Robach.',\\
  `Amy Robach is a journalist.',\\
  `Amy Robach is a news anchor.',\\
  `Andrew Shue and Amy Robach have two daughters together.'] \\

\hline
\end{tabular}
\caption{Output of long-text decomposition. The question is from FactScore-Bio and the model used to generate the answer is Llama-3-8B.}
\label{tab:decomp_sample2}
\end{table*}

\paragraph{Labeling Decomposed Claims} 
Long-form generations, or decomposed claims, typically lack ground truths, essential for assessing the performance of uncertainty estimation. To address this issue, we adopt the methodology named as Search-Augmented Factuality Evaluator (SAFE)\cite{wei2024longformsafe}. SAFEemploys Google Search to retrieve passages related to each claim, then applies reasoning with an LLM to determine whether the claim is supported or unsupported. In our evaluations, we utilize GPT-4o-mini as the LLM for reasoning and consider supported claims as correct and unsupported claims as incorrect. 
We use the original implementation and the default prompts and settings provided by that\footnote{\url{https://github.com/google-deepmind/long-form-factuality/blob/main/eval/safe/rate_atomic_fact.py}}.

\paragraph{Prompts Used in Proposed Strategies} To generate questions within the QG and QAG strategies described in Section~\ref{lfg-exp-des}, we used the prompt provided in Table~\ref{tab:qgen_prompt}. Note that the base model that is used the generate an answer to the main question is also used to generate questions. To generate an answer to the generated question within QAG strategy, we employed the following prompt:
\begin{verbatim}
You are a helpful assistant. Give a single 
claim answer to given question. Don't 
provide any additional information. Just 
answer the question with a brief sentence 
in a single claim. Question: {question} 
Answer: 
\end{verbatim}

\begin{table*}[ht]
\centering
\begin{tabular}{|p{\textwidth}|}
\hline
\textbf{System:} You are an expert assistant skilled at generating focused and contextually relevant questions from claims. Your task is to create a question such that the answer would align closely with the provided claim. To ensure the question is precise and relevant, consider the context provided by the original question. Study the examples below from a variety of topics and follow the same pattern. \\ \\

Original Question: What themes are commonly explored in 20th-century dystopian literature?

Claim: George Orwell's novel 1984 explores the theme of government surveillance. 

Question: What theme does George Orwell's novel 1984 explore?    \\ \\

Original Question: What themes are commonly explored in 20th-century dystopian literature?

Claim: George Orwell's novel 1984 portrays a totalitarian regime that monitors every aspect of citizens' lives. 

Question: How does George Orwell's novel 1984 reflect the theme of totalitarian control, as commonly explored in 20th-century dystopian literature?  \\ \\

Original Question: What themes are commonly explored in 20th-century dystopian literature?

Claim: The novel 1984 is written by George Orwell.

Question: Who has written the novel 1984? \\ \\

Original Question: How has artificial intelligence influenced industries in the 21st century?

Claim: Artificial intelligence enables better decision-making through data analysis.

Question: How does artificial intelligence enhance the decision-making process in modern businesses? \\ \\

Original Question: What factors contributed to the Great Depression, and how did governments respond?

Claim: Stock market speculation contributed to the Great Depression.

Question: Did stock market speculation contribute to the Great Depression?  \\ \\

Original Question: Who is Abraham Lincoln?

Claim: Abraham Lincoln is best known for leading the country through the Civil War.

Question: What is Abraham Lincoln's most significant historical contribution? \\ \\

Original Question: Who is Abraham Lincoln?

Claim: Abraham Lincoln served from 1861 to 1865 as the president of the US.

Question: When did Abraham Lincoln serve as the president of the United States?  \\ \\

Now, follow the pattern demonstrated in the examples to generate a question for the given claim, without adding explanations, introductions, or conversational responses. \\ \\

\textbf{User:} Original question: \{MAIN\_QUESTION\}

Claim: \{CLAIM\}

Question:  \\
\hline
\end{tabular}
\caption{Prompt for question generation used in QG and QAG strategies to adapt UE methods to long-form generation.}
\label{tab:qgen_prompt}
\end{table*}

\paragraph{Dataset Details}
We provide sample questions from the datasets in Table \ref{tab:dataset_samples_lfg}. 
For FactScore-Bio, the total number of claims is 1290 for GPT-4o-mini and 1764 for Llama-3-8B. 
For LongFact-Objects, the total number of claims is also 1049 for GPT-4o-mini and 1780 for Llama-3-8B. 
Note that the same set of questions is used for both models.

\subsection{Computational Budget}
We use 40 GB Nvidia A100 GPUs for all the experiments. We use GPT API to run gpt-4o-mini experiments. The total GPU hours for LLama-3-8b experiments are approximately 800 hours.

\begin{table*}[!htbp]
\centering\centering
\fontsize{10.5}{11}\selectfont
\begin{tabular}{c|l}
\toprule
Dataset & Question \\
\midrule[\heavyrulewidth]
\multirow{5}{*}{{\textbf{FactScore-Bio}}}&
Tell me a bio of Vaira Vīķe-Freiberga. \\
&Tell me a bio of Ji Sung. \\
&Tell me a bio of Baltasar Corrada del Río. \\
&Tell me a bio of Henry Santos. \\
&Tell me a bio of Mike Trivisonno. \\
\midrule
\multirow{5}{*}{{\textbf{LongFact-Objects}}}&
Who is Yoshua Bengio? \\
&What is known about the World Trade Organization? \\
&What took place during the fall of the Berlin Wall in 1989? \\
&What is the gaming company "Blizzard Entertainment"? \\
&How is the United States related to the East Asia Summit (EAS)? \\
\bottomrule
\end{tabular}
\vskip -0.1in
\caption{Sample questions from long-form generation datasets.}
\label{tab:dataset_samples_lfg}
\end{table*}

\end{document}